\documentclass[11pt]{article}
\usepackage{acl}
\usepackage{times}
\usepackage{latexsym}
\usepackage{booktabs}
\usepackage{amsmath}
\usepackage[T1]{fontenc}
\usepackage[utf8]{inputenc}
\usepackage{microtype}
\IfFileExists{placeins.sty}{\usepackage{placeins}}{\providecommand{\FloatBarrier}{}}
\IfFileExists{inconsolata.sty}{\usepackage{inconsolata}}{}
\makeatletter
\IfFileExists{fvextra.sty}{\usepackage{fvextra}}{\usepackage{fancyvrb}\define@key{FV}{breaklines}[true]{}\define@key{FV}{breakanywhere}[true]{}}
\makeatother
\usepackage{graphicx}
\graphicspath{{figures/}}
\usepackage{subcaption}
\usepackage{tabularx}
\usepackage{xcolor}
\usepackage{tikz}
\usetikzlibrary{positioning,arrows.meta,fit,calc,decorations.pathreplacing}

\newcommand{\nParticipants}{46}
\newcommand{\nCohortOne}{20}
\newcommand{\nCohortTwo}{26}
\newcommand{\nItems}{25}
\newcommand{\nChoiceItems}{17}
\newcommand{\nLikertItems}{7}
\newcommand{\nRankingItems}{1}
\newcommand{\nOptionsMin}{3}
\newcommand{\nOptionsMax}{4}

\newcommand{\nInterviewDomains}{10}

\newcommand{\chanceChoiceMin}{0.25}
\newcommand{\chanceChoiceMax}{0.33}
\newcommand{\nStaticQuestions}{4}
\newcommand{\nStaticIncomplete}{6}

\newcommand{\tokFollowupMedianCohortTwo}{532}

\newcommand{\tokStaticMedian}{368}
\newcommand{\tokRatioStaticAdaptive}{0.73}

\newcommand{\nDrawsPerCohort}{10}

\newcommand{\accMajority}{0.407}

\newcommand{\accNoPersona}{0.392}

\newcommand{\accCore}{0.395}

\newcommand{\accFull}{0.400}

\newcommand{\accSummary}{0.388}

\newcommand{\accNoPersonaAltGptFivePointOneCohortOne}{0.428}

\newcommand{\accNoPersonaAltGptFivePointOneCohortTwo}{0.370}

\newcommand{\cFullStaticAccA}{0.402}
\newcommand{\cFullStaticAccB}{0.394}
\newcommand{\cFullStaticDiff}{+0.007}
\newcommand{\cFullStaticCI}{[-0.028, +0.042]}
\newcommand{\cFullStaticP}{0.683}

\newcommand{\cFullStaticD}{0.08}

\newcommand{\cFullStaticPartA}{12}
\newcommand{\cFullStaticPartB}{14}

\newcommand{\cCoreNoPersonaDiff}{+0.003}
\newcommand{\cCoreNoPersonaCI}{[-0.030, +0.035]}
\newcommand{\cCoreNoPersonaP}{0.844}

\newcommand{\cFullCoreDiff}{+0.005}
\newcommand{\cFullCoreCI}{[-0.017, +0.027]}
\newcommand{\cFullCoreP}{0.652}

\newcommand{\cSummaryCoreDiff}{-0.007}
\newcommand{\cSummaryCoreCI}{[-0.043, +0.028]}

\newcommand{\cGroundedAccA}{0.445}
\newcommand{\cGroundedAccB}{0.397}
\newcommand{\cGroundedDiff}{+0.048}
\newcommand{\cGroundedCI}{[-0.035, +0.132]}
\newcommand{\cGroundedP}{0.253}

\newcommand{\gUsedAdaptiveCohortTwoPct}{38}

\newcommand{\gUsedStaticPct}{59}
\newcommand{\gGroundedAccStatic}{0.389}

\newcommand{\gCoreOnlyAccStatic}{0.372}

\newcommand{\nWrongTraces}{453}
\newcommand{\nControlTraces}{60}
\newcommand{\eBShare}{0.850}
\newcommand{\eBPct}{85}

\newcommand{\eBCI}{[0.814, 0.884]}

\newcommand{\eBShareHighConf}{0.904}
\newcommand{\eARetShare}{0.038}

\newcommand{\eARetCI}{[0.021, 0.055]}

\newcommand{\eCIntShare}{0.044}

\newcommand{\eCIntCI}{[0.027, 0.062]}

\newcommand{\eAReasShare}{0.068}

\newcommand{\eAReasN}{31}
\newcommand{\eAReasCI}{[0.044, 0.095]}

\newcommand{\eBRelatedPct}{60}
\newcommand{\eBNoMaterialPct}{40}

\newcommand{\eConfLowPct}{11}
\newcommand{\eFlagInconsistent}{9}
\newcommand{\eCtrlEvidencePct}{63}
\newcommand{\eWrongEvidencePct}{15}
\newcommand{\eValidationProblems}{0}
\newcommand{\nBlindCells}{782}

\newcommand{\fCoveragePct}{52}

\newcommand{\fFullAccStrengthZero}{0.410}
\newcommand{\fStrengthOneN}{287}
\newcommand{\fFullAccStrengthOne}{0.422}
\newcommand{\fStrengthTwoN}{108}
\newcommand{\fFullAccStrengthTwo}{0.454}
\newcommand{\fStrengthThreeN}{9}
\newcommand{\fFullAccStrengthThree}{0.444}

\newcommand{\fPrecisionPct}{35}
\newcommand{\fReaderAcc}{0.180}

\newcommand{\fFullAccYes}{0.431}
\newcommand{\fFullAccNo}{0.410}
\newcommand{\fFullDiff}{+0.021}
\newcommand{\fFullCI}{[-0.047, +0.087]}
\newcommand{\fFullP}{0.547}

\newcommand{\fNoPersonaDiff}{+0.030}

\newcommand{\fMajorityAccYes}{0.483}

\newcommand{\fMajorityDiff}{+0.112}
\newcommand{\fMajorityCI}{[+0.031, +0.199]}
\newcommand{\fMajorityP}{0.002}

\newcommand{\fRichnessFullRho}{+0.02}
\newcommand{\fRichnessFullP}{0.87}

\newcommand{\fRichnessTokensRho}{+0.51}
\newcommand{\fRichnessTokensP}{\ensuremath{<}0.01}

\newcommand{\fValidationProblems}{0}

\newcommand{\mMbtiCore}{0.326}

\newcommand{\mMbtiFull}{0.543}

\newcommand{\mMbtiDiff}{+0.217}
\newcommand{\mMbtiCI}{[+0.043, +0.391]}
\newcommand{\mMbtiP}{0.032}
\newcommand{\mMbtiUp}{14}
\newcommand{\mMbtiDown}{4}
\newcommand{\mBigFiveDiff}{+0.013}

\newcommand{\mDecisionDiff}{+0.005}

\newcommand{\hWithinCos}{0.61}
\newcommand{\hBetweenCos}{0.34}

\newcommand{\hNQuestions}{440}
\newcommand{\hNParticipants}{44}
\newcommand{\hParticipantsPerfect}{41}
\newcommand{\iTokensFullRho}{+0.07}
\newcommand{\iTokensFullP}{0.60}

\newcommand{\iTokensCoreRho}{-0.08}
\newcommand{\iTokensCoreP}{0.56}

\newcommand{\iTokensSummaryRho}{-0.07}
\newcommand{\iTokensSummaryP}{0.63}

\newcommand{\iFollowupTokensLiftRhoWithin}{+0.30}
\newcommand{\iFollowupTokensLiftPWithin}{0.05}

\newcommand{\jFullGrounded}{0.445}

\newcommand{\jCoreGrounded}{0.409}

\newcommand{\jNoPersonaDiff}{+0.027}

\newcommand{\jMajorityDiff}{+0.073}
\newcommand{\jMajorityCI}{[+0.001, +0.151]}

\newcommand{\jTransGroundedImproved}{32}
\newcommand{\jTransGroundedWorsened}{21}
\newcommand{\jTransGroundedP}{0.169}

\newcommand{\xModelName}{DeepSeek-V4 Pro}
\newcommand{\xFullAcc}{0.357}
\newcommand{\xFullCI}{[0.331, 0.382]}

\newcommand{\xCoreAcc}{0.350}
\newcommand{\xCoreCI}{[0.327, 0.374]}

\newcommand{\xNoPersonaAcc}{0.372}
\newcommand{\xNoPersonaCI}{[0.338, 0.404]}

\newcommand{\xNoPersonaDraws}{10}
\newcommand{\xFullCoreDiff}{+0.007}
\newcommand{\xFullCoreCI}{[-0.014, +0.027]}
\newcommand{\xFullCoreP}{0.542}

\newcommand{\xFullMajorityDiff}{-0.050}
\newcommand{\xFullMajorityCI}{[-0.082, -0.017]}
\newcommand{\xFullMajorityP}{0.006}

\newcommand{\xFullNoPersonaDiff}{-0.015}
\newcommand{\xFullNoPersonaCI}{[-0.049, +0.020]}
\newcommand{\xFullNoPersonaP}{0.396}

\newcommand{\xFollowupSharePct}{51}

\newcommand{\xQuotesFoundPct}{97}
\newcommand{\xQuotes}{1525}

\newcommand{\xLabelAgreeBinary}{0.936}
\newcommand{\xLabelKappaBinary}{0.87}

\newcommand{\xGptFollowupSharePct}{38}
\newcommand{\xGroundedAcc}{0.379}
\newcommand{\xCoreOnlyAcc}{0.347}
\newcommand{\xGroundedDiff}{+0.032}
\newcommand{\xGroundedCI}{[-0.044, +0.107]}
\newcommand{\xGroundedP}{0.403}

\newcommand{\xGroundedDiffV}{+0.027}
\newcommand{\xGroundedCIV}{[-0.039, +0.092]}
\newcommand{\xGroundedPV}{0.477}

\newcommand{\xSameGptFullDiff}{-0.018}
\newcommand{\xSameGptFullCI}{[-0.081, +0.042]}

\newcommand{\xSameMajorityDiff}{-0.003}
\newcommand{\xSameMajorityCI}{[-0.075, +0.068]}

\newcommand{\xTransGroundedImproved}{46}
\newcommand{\xTransGroundedWorsened}{29}
\newcommand{\xTransGroundedP}{0.064}
\newcommand{\xPairedGroundedFull}{0.379}
\newcommand{\xPairedGroundedCore}{0.336}
\newcommand{\xPairedGroundedGain}{+0.043}
\newcommand{\xPairedGroundedGainCI}{[+0.007, +0.079]}

\newcommand{\xPairedCoreOnlyGain}{-0.031}
\newcommand{\xPairedCoreOnlyGainCI}{[-0.064, +0.003]}

\newcommand{\xOverlapKappa}{0.33}

\newcommand{\xOverlapN}{782}
\newcommand{\hvTraces}{120}
\newcommand{\hvOverlap}{60}
\newcommand{\hvAnnotators}{3}
\newcommand{\hvIRAPct}{95}
\newcommand{\hvAgreePct}{87.9}

\newcommand{\qAgreeFull}{0.513}
\newcommand{\qConsistentFull}{0.483}

\newcommand{\qHumanConsistent}{0.326}
\newcommand{\vChatgptAcc}{0.388}
\newcommand{\vCoreAcc}{0.428}
\newcommand{\vFullAcc}{0.628}

\newcommand{\vNoPersonaAcc}{0.500}
\newcommand{\vMajorityAcc}{0.360}
\newcommand{\vRank}{1}
\newcommand{\vRankOf}{20}
\newcommand{\vSameAsFull}{16}

\newcommand{\kCodexBlindAgree}{0.81}
\newcommand{\kCodexBlindKappa}{0.63}

\newcommand{\kCodexBlindCoverage}{0.459}
\newcommand{\kCodexBlindPrecision}{0.343}
\newcommand{\kCodexBlindFullDiff}{+0.051}
\newcommand{\kCodexBlindFullP}{0.153}
\newcommand{\kCodexInformedAgree}{0.91}
\newcommand{\kCodexInformedKappa}{0.58}
\newcommand{\kCodexInformedKappaB}{0.66}
\newcommand{\kCodexInformedBShare}{0.905}

\newcommand{\kCodexBShareOfARetPct}{47}
\newcommand{\kCodexBShareOfCIntPct}{45}
\newcommand{\kCodexBShareOfAReasPct}{39}

\newcommand{\coreten}{Core-10}
\newcommand{\full}{Full}
\newcommand{\static}{Static}
\newcommand{\summ}{Summary}
\newcommand{\nopersona}{No-persona}
\newcommand{\majority}{Majority}

\title{What a One-Hour Interview Does Not Tell an LLM About You:\\Interviews Elicit Traits, Not Decision Policies}

\author{Ruoxi Su \\ Independent Researcher \\ {\small\texttt{letitia.su@gmail.com}} \And
  Yuhan Liu \\ University of Cambridge \\ {\small\texttt{yl972@cantab.ac.uk}} \And
  Jingyu Hu \\ Independent Researcher \\ {\small\texttt{jingyu.hu@alumni.utoronto.ca}}}

\begin{document}
\maketitle
\begin{abstract}
Large language models are increasingly asked to stand in for particular people, and interviewing the person is the natural way to tell a model who they are. We test what a one-hour LLM-led adaptive interview tells a model about an individual's value-laden decisions. \nParticipants{} adults completed such an interview and then answered \nItems{} dilemma questions, which are the ground truth. We predict their answers from different parts of the transcript, compare adaptive follow-up questions with fixed ones on the same participants, and benchmark every variant against a model given no information about the person and against the majority answer. No part of the interview is reliably better than either baseline, and adaptive follow-ups are no better than fixed ones. Two audits of the transcripts show why. For \eBPct\% of the errors an auditor finds no evidence that distinguishes the person's choice from the alternatives, and where a transcript does contain related evidence, the option it favours matches the person's own answer less often than the majority answer does. The same transcripts do improve recovery of the person's MBTI type. Interviews of this kind elicit traits, not decision policies. We discuss what decision-aware elicitation would require and why persona agents should not be delegated value-laden decisions, and we release the consenting participants' transcripts with all predictions.
\end{abstract}

\section{Introduction}
\label{sec:intro}

Large language models (LLMs) are increasingly used as stand-ins for particular people: to answer a survey as a given respondent, to rehearse a decision before the person makes it, or to act on their behalf as a personal agent \citep{park2024generative,li2025hugagent,deng2026aimoderated}. Few people have years of recorded behaviour to hand a model, so the natural way to tell it about a person is to interview them. Agents built from two-hour AI-led interviews reproduce held-out survey answers at a large fraction of each person's own consistency on retest \citep{park2024generative}, and lighter interviews are proposed for market research \citep{deng2026aimoderated,wang2026interview}.

What such an interview tells a model about a person's \emph{decisions} is less clear. Persona-conditioned models reproduce population statistics but add little respondent-specific variance beyond demographics \citep{peng2025funhouse,ahn2026itemmean,tadaymorocho2026reliability}, and a pre-registered comparison found twins built from adaptive AI interviews no more accurate than twins built from static questionnaires \citep{deng2026aimoderated}. These results are outcome-level: they show that information is lost between the person and the prediction, not where. The transcript may lack the relevant evidence; the model may fail to retrieve evidence that is there; it may fail to integrate evidence spread over several turns; or it may retrieve the right passage and still reason to the wrong answer. Each failure calls for a different remedy, and to our knowledge no study of interview-based simulation has separated them.

We separate them for \emph{value-laden decisions}: choices among courses of action that trade off competing values or obligations and have no correct answer \citep{haidt2001emotional,hendrycks2021aligning}. \nParticipants{} adults completed a one-hour LLM-led interview (ten core questions over fixed psychological domains, three to five adaptive follow-ups, a summary) and then answered \nItems{} dilemma questions themselves; their answers are the ground truth. We predict those answers from different parts of the transcript, add a within-subject comparison in which the same people also answered four fixed follow-up questions, and benchmark every input against two baselines: the same model given nothing about the person, and the majority answer of the other participants. We then audit the transcripts twice. An informed audit assigns every wrong prediction to one of the four failures above. A blind audit that never sees answers or predictions records, for every participant and question, whether the transcript contains decision-relevant evidence and which option it favours.

No part of the interview is reliably better than either baseline, and adaptive follow-ups do no better than fixed ones. The audits locate the loss in the transcripts. For \eBPct\% of the errors the auditor finds no evidence that distinguishes the person's choice from the alternatives, and where the blind audit finds related evidence, the option it favours matches the person's own answer less often than the majority answer does. This is a \emph{diagnosticity gap}: the interviews elicit dispositions that bear on the dilemmas but not the policies that decide them. The same transcripts raise MBTI type recovery from \mMbtiCore{} to \mMbtiFull{}. Interviews of this kind elicit traits, not decision policies.

Our contributions are (i) a controlled test of interview-based decision simulation, with the same participants, the same core questions, adaptive versus fixed follow-ups and two baselines; (ii) a two-level audit that locates where the information is lost, an informed error decomposition and a blind evidence audit, each re-labelled by a second annotator from a different model family; (iii) a dissociation on the same transcripts between personality recovery, which improves with adaptive follow-ups, and decision prediction, which does not; and (iv) a dataset of paired interview transcripts, \nItems{} decisions and personality self-reports for \nParticipants{} people, with all predictions (\S\ref{sec:data}).

\begin{figure*}[t]
\centering
\footnotesize
\resizebox{\textwidth}{!}{%
\begin{tikzpicture}[
  node distance=4pt and 10pt,
  box/.style={draw, rounded corners=2pt, align=center, inner sep=3pt, minimum height=5mm, font=\footnotesize, text width=3.4cm},
  llm/.style={box, fill=blue!9},
  human/.style={box, fill=orange!14},
  data/.style={box, fill=green!12},
  pred/.style={box, fill=purple!8, text width=4.1cm},
  panel/.style={draw, rounded corners=3pt, inner sep=6pt, dashed, gray!70},
  ptitle/.style={font=\footnotesize\bfseries, anchor=south west, inner sep=1pt},
  tag/.style={font=\scriptsize, text=purple!70!black},
  brace/.style={decorate, decoration={brace, amplitude=3pt}, thick, purple!70!black},
  arr/.style={-{Latex[length=1.5mm]}, thick}]
\node[llm] (q10) {Interviewer LLM asks \nInterviewDomains{} core questions,\\one per fixed psychological domain};
\node[human, below=of q10] (a10) {Participant answers};
\node[llm, below=of a10] (fu) {Interviewer asks 3--5 adaptive follow-ups,\\quoting the participant's answers};
\node[human, below=of fu] (afu) {Participant answers};
\node[llm, below=of afu] (sum) {Interviewer writes a per-domain summary};
\foreach \a/\b in {q10/a10, a10/fu, fu/afu, afu/sum} \draw[arr] (\a) -- (\b);
\draw[brace] ([xshift=3pt]q10.north east) -- ([xshift=3pt]a10.south east) node[tag, midway, right=4pt] {\coreten};
\draw[brace] ([xshift=30pt]q10.north east) -- ([xshift=30pt]afu.south east) node[tag, midway, right=4pt] {\full};
\draw[brace] ([xshift=3pt]sum.north east) -- ([xshift=3pt]sum.south east) node[tag, midway, right=4pt] {\summ};
\node[panel, fit={(q10) (sum) ([xshift=56pt]afu.east)}] (pA) {};
\node[ptitle] at (pA.north west) {(a) Interview (LLM-led, about 1 h)};
\node[human, right=52pt of q10.north east, anchor=north west, text width=3.2cm] (items) {\nItems{} dilemma questions\\(\nChoiceItems{} choice, \nLikertItems{} Likert, \nRankingItems{} ranking)};
\node[human, below=of items, text width=3.2cm] (bfi) {BFI-44 (Big Five)};
\node[human, below=of bfi, text width=3.2cm] (mbti) {MBTI self-report};
\node[human, below=of mbti, text width=3.2cm, fill=gray!12] (fixed) {\nStaticQuestions{} fixed follow-up questions,\\identical for all (cohort 2)};
\foreach \a/\b in {items/bfi, bfi/mbti, mbti/fixed} \draw[arr] (\a) -- (\b);
\draw[brace] ([xshift=3pt]fixed.north east) -- ([xshift=3pt]fixed.south east) node[tag, midway, right=4pt, align=left] {\static{} =\\core Q\&A\\+ these};
\node[panel, fit={(items) (fixed) ([xshift=40pt]fixed.east)}] (pB) {};
\node[ptitle] at (pB.north west) {(b) Questionnaire};
\node[pred, right=56pt of items.north east, anchor=north west] (inp) {Predictor LLM receives one \emph{input regime}: \coreten, \full, \static{} or \summ};
\node[pred, below=of inp] (out) {Answers the \nItems{} questions as the participant, with a \emph{trace}: rationale, verbatim quotations, confidence; a second call labels reasoning type and quotation location};
\node[pred, below=of out] (per) {Predicts MBTI (two types) and Big Five scores};
\node[data, below=of per, text width=4.1cm] (cmp) {Compared with the participant's own questionnaire answers};
\draw[arr] (inp) -- (out); \draw[arr] (out) -- (per); \draw[arr] (per) -- (cmp);
\node[panel, fit=(inp) (cmp)] (pC) {};
\node[ptitle] at (pC.north west) {(c) Simulation (GPT-5)};
\draw[arr, gray!60!black] (pA.east |- fu.west) -- node[above, font=\scriptsize, text=black] {transcript} (pB.west |- fu.west);
\draw[arr, gray!60!black] (pB.east |- cmp.west) -- node[above, font=\scriptsize, text=black] {ground truth} (pC.west |- cmp.west);
\end{tikzpicture}}
\caption{Study procedure. (a) An LLM interviews each participant in three stages; braces mark the parts of the transcript that form the \coreten, \full{} and \summ{} regimes. (b) The participant answers the \nItems{} dilemma questions (the ground truth), the BFI-44 and an MBTI self-report; cohort 2 also answers four fixed follow-ups, which replace the adaptive ones in the \static{} regime. (c) A predictor LLM receives one regime and answers the same questions and inventories, scored against the participant's own answers. Baselines (\S\ref{sec:regimes}) and audits (\S\ref{sec:audits}) involve no participant.}
\label{fig:pipeline}
\end{figure*}

\section{Related Work}
\label{sec:related}

\paragraph{Simulating individuals.}
Agents built from two-hour AI-led interviews reproduce held-out survey answers better than agents built from demographics alone \citep{park2024generative}, and large panels supply retest ceilings for scoring them \citep{toubia2025twin}. Later work is sobering about individuation: twins built from full personas are no more accurate than demographics-only twins and are under-dispersed \citep{peng2025funhouse}; richer persona data mostly reproduce question means rather than respondent-specific variance \citep{ahn2026itemmean}; static personas yield near-chance exact match on opinion questions \citep{tadaymorocho2026reliability}; and interview-informed agents are ``identity-imprecise'' in product discovery \citep{wang2026interview}. Raw history beats summaries, structure matters more than volume \citep{kinzinger2026synthetic,ye2026beyond,pourtaheri2026behavioral}, and persona completion can be focused on the query at hand \citep{su2026queryfocused}. These are outcome-level findings; none locates where in the chain from elicitation to prediction the information is lost. Appendix~\ref{app:related-table} compares the designs.

\paragraph{Adaptive versus fixed questioning.}
The closest test is a pre-registered between-subjects comparison of AI-moderated adaptive interviews with static questionnaires, which found no difference in twin accuracy on marketing ratings \citep{deng2026aimoderated}; its static arm had no follow-ups and the arms were not matched on information budget. Adaptive querying wins only at small question budgets \citep{wang2026adaptive}, and in survey methodology AI-led probing raises data quality and coverage without a demonstrated gain in prediction \citep{chopra2026qualitative,austin2026experimental,panfilova2026interviewer}. LLM-led preference elicitation brings no significant gain in the moral domain \citep{li2023gate}, and an LLM-led interview paired with individual belief prediction shows non-monotonic returns to context \citep{li2025hugagent}. Our comparison is within subject, with the same core questions, a comparable number of follow-ups, reported token budgets and decision outcomes.

\paragraph{Personas, evidence and faithfulness.}
Persona prompting changes reasoning style at the population level \citep{nighojkar2025personalities} but misses individual trait levels \citep{petrov2024limited}; personas built from digital traces or years of public comments do better \citep{li2026persona,lee2025solar}. None of this work asks whether the elicited material is diagnostic of the target decisions. Total Simulated Survey Error separates persona-construction from response-generation error but has no retrieval or integration node \citep{sen2026ts2e}. Citation correctness is not faithfulness \citep{wallat2024correctness,cohenwang2024contextcite,deng2026factchecking}, and persona-conditioned models fabricate profile claims \citep{sun2026mirage,ren2026neutrality}, so we do not rely on the predictor's own evidence labels. LLM judges favour their own generations \citep{panickssery2024evaluators}, so our auditors come from a different model family than the predictor.

\section{Method}
\label{sec:method}

Figure~\ref{fig:pipeline} shows the procedure. An LLM interviews each participant, who then answers the dilemma questions and personality inventories that are the ground truth (\S\ref{sec:protocol} to \ref{sec:items}). A predictor LLM answers the same questions under several \emph{input regimes}, that is, different parts of the participant's transcript, and we compare its accuracy with two baselines that use no interview at all (\S\ref{sec:regimes} and \ref{sec:stats}). Two audits then examine the transcripts and the predictor's reasoning traces to locate where the information needed for a correct prediction is lost (\S\ref{sec:audits}).

\subsection{Interview protocol}
\label{sec:protocol}
The interviewer is an LLM driven by a single meta-prompt (Appendix~\ref{app:meta-prompt}) in three stages. In stage 1 it generates \nInterviewDomains{} open-ended core questions, one for each of \nInterviewDomains{} psychological domains fixed in the prompt (Appendix~\ref{app:meta-prompt} lists them with their psychological basis). The wording is generated afresh for each participant so that the questions invite concrete stories rather than self-ratings; the domains do not vary, and an embedding analysis confirms that the generated questions are stable across participants (within-domain cosine \hWithinCos{} against \hBetweenCos{} between domains; Appendix~\ref{app:consistency}). In stage 2, after reading the ten answers, the interviewer asks three to five adaptive follow-up questions that target unclear, inconsistent, emotionally charged or characteristic material and quote the participant's own words. In stage 3 it writes a per-domain summary of the participant. Participants typed their answers through a survey form; a session took about an hour. They were recruited in two waves, which we call cohorts 1 and 2 (\S\ref{sec:participants}); the interviewer, under the same meta-prompt, was DeepSeek-R1 \citep{deepseek_r1_2025} in deep-thinking mode for cohort 1 and DeepSeek-V4 in deep-thinking mode for cohort 2, since R1 had been withdrawn from the API in between.

\paragraph{Fixed follow-up control.}
To separate the effect of \emph{adaptive} follow-ups from that of asking more questions, every participant in cohort 2 also answered \nStaticQuestions{} fixed follow-up questions, identical for everyone and written by the authors to probe the kinds of material adaptive follow-ups typically target, such as a forced compromise or a rule and an instance of breaking it (Appendix~\ref{app:static}). They were answered after the dilemma questions, so they could not prime the decisions, although the decisions could in principle prime them. Fixed question-and-answer blocks are somewhat shorter than adaptive ones (median \tokStaticMedian{} against \tokFollowupMedianCohortTwo{} tokens; median per-participant ratio \tokRatioStaticAdaptive), so the control has a slightly smaller budget. For \nStaticIncomplete{} participants one fixed question drew a non-informative answer (typically that no such experience came to mind), which was dropped for them.

\subsection{Participants and ground truth}
\label{sec:participants}
\nParticipants{} adults took part. Cohort 1 (\nCohortOne) was recruited in October 2025 through campus networks in North America and cohort 2 (\nCohortTwo) in September 2026 through online media, with some further participants from North America. Aggregate demographics are in Appendix~\ref{app:demographics}. After the interview, each participant answered the \nItems{} dilemma questions described next; their own answers are the ground truth throughout. They also completed the Big Five Inventory \citep[BFI-44; scored 1 to 40 per trait]{john1999bfi} and reported their MBTI type \citep{myers1998mbti}, giving two types when unsure. All participants consented to research use and public release of their de-identified data (Ethical Considerations).

\subsection{Decision questions}
\label{sec:items}
The \nItems{} questions are short scenarios in which the respondent must choose a course of action that trades off competing values or obligations; none has a correct answer. \nChoiceItems{} are single-choice questions with \nOptionsMin{} to \nOptionsMax{} options, \nLikertItems{} are five-point Likert questions, and one (Q17) asks for a ranking of five options. Each question was adapted from a named paradigm in moral or social psychology (for example trolley variants, dictator allocations and Schwartz value conflicts) and set in an everyday scene; Appendix~\ref{app:items} lists the questions with their paradigms, chance levels and thematic domains.

\subsection{Input regimes and baselines}
\label{sec:regimes}
An input regime is the part of a participant's interview the predictor sees. \coreten{} gives the ten core questions with their answers. \full{} gives the complete transcript: the core block followed by the adaptive follow-ups and their answers. \static{} (cohort 2 only) gives the core block followed by the four fixed follow-ups and their answers in place of the adaptive ones. \summ{} gives only the interviewer's stage-3 summary.

The predictor was GPT-5 \citep{openai_gpt5_2025} at reasoning effort medium throughout; a replication with an open-weight predictor, \xModelName{}, is in Appendix~\ref{app:crossmodel}. For \coreten, \full{} and \static{} one prompt (Appendix~\ref{app:pred-prompt}) instructs the model to answer as the interviewee from the supplied persona information alone and to return, for each question, a short rationale, the predicted answer, up to two verbatim supporting quotations and a confidence label; a second, annotation-only call labels each prediction's reasoning type and the block (core or follow-up) its quotations come from. We call this structured output the prediction's \emph{trace}; all decision accuracies below come from the runs that produced traces. The \summ{} regime uses a prompt without the quotation fields, since a summary has no turns to cite (Appendix~\ref{app:summary-prompt}; cohort-1 \summ{} predictions are from a prediction-only run). Personality is predicted in a separate prediction-only run per regime (two candidate MBTI types in order of likelihood; a score per Big Five trait).

Two baselines use no interview. \nopersona{} gives the same predictor the \nItems{} questions with no information about any person and asks how an ordinary respondent would answer (Appendix~\ref{app:nopersona-prompt}). We took \nDrawsPerCohort{} complete draws per cohort wording, with the same model and reasoning effort as the persona runs, and score each participant against the average over draws, so question scores can be fractional. \majority{} assigns each participant, for each question, the modal answer of all other participants (the leave-one-out mode; the modal permutation for Q17 and, for Likert ties, the mode nearest the leave-one-out median). It is the best constant guess from the sample and the reference point recommended by recent twin studies \citep{peng2025funhouse,bojic2026persona,ahn2026itemmean}.

\subsection{Metrics and statistics}
\label{sec:stats}
A single-choice or Likert question scores 1 if the predicted answer equals the participant's own answer and 0 otherwise; Q17 scores the share of the ten option pairs whose relative order the prediction reproduces. Overall accuracy is the mean question score over the \nItems{} questions and all participants; per-type accuracies and a one-point tolerance for Likert questions are in Appendix~\ref{app:types}. MBTI recovery is hit@2, whether either predicted type is in the participant's self-reported set; Big Five scores are binarised at 21 and matched per trait.

Five contrasts were specified before the data were analysed: (1) \full{} versus \static{} within cohort 2, the primary test of adaptive against fixed follow-ups; (2) \coreten{} versus \nopersona{} (do the core answers add anything over nothing?); (3) \full{} versus \coreten{} (do the follow-ups add anything over the core answers?); (4) \summ{} versus \coreten{} (does a distilled summary beat the raw core answers?); and (5) within \full{} on the categorical questions, traces that cite follow-up evidence versus traces that cite only core evidence (does the predictor do better when it uses the follow-ups?). \full{} versus \nopersona{} and \full{} versus \summ{} are in Appendix~\ref{app:contrasts} with the per-cohort contrasts. Contrasts 1 to 4 are paired at the participant level and tested with a sign-flip permutation test on participant means (10,000 permutations); contrast 5 shuffles the grounding label within participant. We report Holm-adjusted $p$-values over the family of five, participant-bootstrap 95\% confidence intervals of each difference, Cohen's $d$ on participant means and Cohen's $h$ on question-level proportions. Per-cohort versions are in Appendix~\ref{app:contrasts}.

\subsection{Two audits of the transcripts}
\label{sec:audits}
\begin{figure}[t]
\centering
\footnotesize
\resizebox{\linewidth}{!}{%
\begin{tikzpicture}[
  node distance=5pt and 6pt,
  q/.style={draw, rounded corners=2pt, align=left, inner sep=3pt, text width=4.35cm, fill=blue!6},
  leaf/.style={draw, align=center, inner sep=3pt, text width=2.05cm, fill=red!9},
  lab/.style={font=\scriptsize, inner sep=1pt, fill=white},
  arr/.style={-{Latex[length=1.4mm]}}]
\node[q] (q1) {\textbf{1.}~Does the transcript contain evidence that \emph{discriminates} the participant's option from the others? (Evidence that supports the predicted answer as strongly as the true one does not count.)};
\node[leaf, left=of q1] (B) {\textbf{B} elicitation gap\\\eBShare{} \eBCI};
\node[q, below=of q1] (q2) {\textbf{2.}~Does the predictor's cited excerpt cover that evidence?};
\node[leaf, left=of q2] (Ar) {\textbf{A-retrieval}\\\eARetShare{} \eARetCI};
\node[q, below=of q2] (q3) {\textbf{3.}~Was the missed value or constraint spread over two or more interview turns?};
\node[leaf, left=of q3] (C) {\textbf{C} integration\\\eCIntShare{} \eCIntCI};
\node[leaf, below=of q3, text width=4.35cm] (Ae) {\textbf{A-reasoning}: apt excerpt, wrong inference\quad \eAReasShare{} \eAReasCI};
\draw[arr] (q1) -- node[lab] {no} (B);
\draw[arr] (q1) -- node[lab] {yes} (q2);
\draw[arr] (q2) -- node[lab] {no} (Ar);
\draw[arr] (q2) -- node[lab] {yes} (q3);
\draw[arr] (q3) -- node[lab] {yes} (C);
\draw[arr] (q3) -- node[lab] {no} (Ae);
\end{tikzpicture}}
\caption{Informed decision tree for every wrong \full{} prediction on the categorical questions; leaves give the share of the \nWrongTraces{} errors (participant-clustered bootstrap 95\% CIs). The blind audit asks only question~1, for every cell, without seeing answers or predictions.}
\label{fig:audits}
\end{figure}

Aggregate accuracy cannot say whether a wrong prediction reflects a transcript that lacked the relevant information or a model that failed to use it. The informed audit classifies each error, which requires seeing the answer, the prediction and the trace; the blind audit measures what the transcripts contain without that knowledge, free of hindsight. We audit every cell, one participant and one question, on the \nChoiceItems{} categorical questions (\nBlindCells{} cells; Likert and ranking questions are excluded because their errors are graded). Both audits were performed by Claude Opus 5.5, a model from a different family than the predictor, since LLM judges tend to favour outputs of their own family \citep{panickssery2024evaluators}, following written protocols (Appendix~\ref{app:e11-prompt} and \ref{app:e15-prompt}) that require every claim about the transcript to be backed by a verbatim quotation; quotations were checked programmatically against the transcripts (\eValidationProblems{} and \fValidationProblems{} violations respectively).

\paragraph{Informed error decomposition.}
For each of the \nWrongTraces{} wrong \full{} predictions, the auditor saw the transcript, the question, the participant's answer, the prediction and its trace, and applied the decision tree in Figure~\ref{fig:audits}. Step 1 asks whether the transcript contains \emph{option-discriminating} evidence, a passage that would materially shift a careful reader's probability for at least one option; passages that support the predicted answer as strongly as the true one do not count, and an answer that contradicts values or constraints the participant states is flagged as unpredictable. A ``no'' is an \textbf{elicitation gap} (B). Step 2 asks whether the predictor's cited excerpt covers that evidence; a ``no'' is a \textbf{retrieval} failure (A-retrieval). Step 3 separates \textbf{integration} failures (C), where the missed value or constraint was spread over two or more interview turns, from \textbf{reasoning} failures (A-reasoning), where a single apt excerpt was cited and the inference from it was wrong. The auditor also applied step 1 to \nControlTraces{} correct predictions; labels were produced in batches of at most eight participants (batch-level shares in Appendix~\ref{app:audit-detail}).

\paragraph{Blind evidence audit.}
Step 1 of the informed audit is made with knowledge of the outcome and is open to hindsight. The blind audit removes it: for every one of the \nBlindCells{} cells the auditor saw only the transcript and the question, never an answer or a prediction, and recorded whether decision-relevant evidence exists, its strength (0 none, 1 an indirect disposition, 2 a clearly applicable value or habit, 3 an explicit statement about this kind of situation), the option or options it favours (the first being the one the auditor would choose from the transcript alone), and the supporting quotations. Because it is independent of correctness, this audit measures what the transcripts contain; crossing it with each regime's correctness then tells us whether the presence of evidence makes any predictor more accurate.

\paragraph{Second annotator.}
To check that the labels are not an artefact of one model family, a second annotator (an OpenAI GPT-6-family model run through Codex, fresh context per participant) relabelled every blind cell and every audited trace from the same packets and protocols, blind pass first and without seeing the first annotator's labels; agreement is reported with the results (\S\ref{sec:results-audit}) and in Appendix~\ref{app:audit-detail}.

\section{Results}
\label{sec:results}

\subsection{No part of the interview beats the baselines}
\label{sec:results-baselines}
\begin{table}[htbp]
\centering
\footnotesize
\setlength{\tabcolsep}{3pt}
\begin{tabular}{lcccc}
\toprule
Condition & $N$ & Acc & 95\% CI & $\Delta$ \\
\midrule
Majority (LOO) & 46 & 0.407 & [0.381, 0.433] & +0.015 \\
No-persona & 46 & 0.392 & [0.356, 0.425] & -- \\
Core-10 & 46 & 0.395 & [0.365, 0.429] & +0.003 \\
Full & 46 & 0.400 & [0.373, 0.431] & +0.008 \\
Summary & 46 & 0.388 & [0.357, 0.421] & -0.004 \\
Static (cohort 2 only) & 26 & 0.394 & [0.348, 0.439] & -- \\
\bottomrule
\end{tabular}
\caption{Overall accuracy on the 25 dilemma questions ($N$=46; Static: cohort 2 only, $N$=26, no $\Delta$). Participant-bootstrap 95\% CIs; $\Delta$ = difference from No-persona, the same model given no information about the person (mean over draws).}
\label{tab:table1}
\end{table}

\begin{table*}[tbp]
\centering
\footnotesize
\setlength{\tabcolsep}{3pt}
\begin{tabular}{lccccccc}
\toprule
Contrast & $N$ & Acc A & Acc B & A$-$B [95\% CI] & $p$ & $p_{\mathrm{Holm}}$ & Effect \\
\midrule
1. Full vs Static (cohort 2, primary) & 26 & 0.402 & 0.394 & +0.007 [-0.028, +0.042] & 0.683 & 1.000 & $d$=0.08, $h$=0.02 \\
2. Core-10 vs No-persona (pooled) & 46 & 0.395 & 0.392 & +0.003 [-0.030, +0.035] & 0.844 & 1.000 & $d$=0.03, $h$=0.01 \\
3. Full vs Core-10 (pooled) & 46 & 0.400 & 0.395 & +0.005 [-0.017, +0.027] & 0.652 & 1.000 & $d$=0.07, $h$=0.01 \\
4. Summary vs Core-10 & 46 & 0.388 & 0.395 & -0.007 [-0.043, +0.028] & 0.711 & 1.000 & $d$=-0.06, $h$=-0.01 \\
5. Grounded vs core-only (Full, cat., pooled) & 46 & 0.445 & 0.397 & +0.048 [-0.035, +0.132] & 0.253 & 1.000 & $h$=0.10 \\
\bottomrule
\end{tabular}
\caption{The five pre-specified contrasts. A$-$B: difference in overall accuracy (contrasts 1--4: 25 questions, participant-level paired) or in accuracy on the categorical questions (contrast 5: question-level, follow-up-grounded vs core-only traces). Participant-clustered permutation $p$ (10{,}000 permutations; sign-flip for paired contrasts, within-participant label shuffle for contrast 5), Holm-adjusted $p$ over the family, participant-bootstrap 95\% CI of the difference, Cohen's $d$ (participant level) and $h$ (question-level proportions).}
\label{tab:e7}
\end{table*}

No input regime is reliably better than either baseline (Table~\ref{tab:table1}; pre-specified contrasts in Table~\ref{tab:e7}). The two baselines bracket the interview regimes: the majority answer reaches \accMajority{} and the model given nothing about the person \accNoPersona, while \coreten, \full{} and \summ{} reach \accCore, \accFull{} and \accSummary. Conditioning on the core answers adds nothing over knowing nothing about the person (\coreten{} $-$ \nopersona{} $=$ \cCoreNoPersonaDiff, 95\% CI \cCoreNoPersonaCI, $p = \cCoreNoPersonaP$), adding the adaptive follow-ups adds nothing over the core answers (\cFullCoreDiff, \cFullCoreCI, $p = \cFullCoreP$), and the summary does slightly worse (\cSummaryCoreDiff, \cSummaryCoreCI). Every Holm-adjusted $p$ equals 1, and the confidence intervals exclude any improvement larger than about three accuracy points. Breakdowns by cohort and question type change nothing (Appendix~\ref{app:types}).

\subsection{Adaptive follow-ups are not better than fixed ones}
\label{sec:results-static}
The primary contrast, within subject, shows no advantage for adaptive follow-ups. For the \nCohortTwo{} participants who answered both kinds, the transcript with adaptive follow-ups yields accuracy \cFullStaticAccA{} and the transcript with the four fixed follow-ups \cFullStaticAccB{} (difference \cFullStaticDiff, 95\% CI \cFullStaticCI, $p = \cFullStaticP$, $d = \cFullStaticD$). The predictor used the fixed material: it cited fixed-question evidence in \gUsedStaticPct\% of its traces on the categorical questions, against \gUsedAdaptiveCohortTwoPct\% for adaptive follow-up evidence in the same participants, and traces grounded in fixed-question evidence were no more accurate than traces grounded in the core answers alone (\gGroundedAccStatic{} against \gCoreOnlyAccStatic; Appendix~\ref{app:contrasts}).

Nor do adaptive follow-ups help when the predictor uses them. Traces that cite follow-up evidence are somewhat more accurate than traces that cite only core evidence (\cGroundedAccA{} against \cGroundedAccB), but the difference is not reliable once inference respects the clustering of questions within participants (\cGroundedDiff, 95\% CI \cGroundedCI, $p = \cGroundedP$), and it is not a property of the follow-ups: the cells on which the predictor cites follow-ups are easier for every predictor, including ones that never saw a follow-up. On the same cells, the majority answer is \jMajorityDiff{} more accurate (95\% CI \jMajorityCI) and the zero-context model \jNoPersonaDiff{} (Appendix~\ref{app:contrasts}, Table~\ref{tab:e17}), so the gap reflects which questions attract citations, not what the citations add. Among the cited cells, the follow-ups turned \jTransGroundedImproved{} wrong Core-10 predictions right and \jTransGroundedWorsened{} right ones wrong ($p = \jTransGroundedP$, sign test). This comparison also rests on the predictor's own report of what it used (human-verified on cohort 1; Appendix~\ref{app:audit-detail}); the audits measure what the transcripts contain. An open-weight predictor shows the same null overall, with a small gain confined to its cited cells (Appendix~\ref{app:crossmodel}).

\subsection{Where the information is lost}
\label{sec:results-audit}
\begin{table}[htbp]
\centering
\small
\begin{tabular}{lccc}
\toprule
Bottleneck & Share & 95\% CI & $n$ \\
\midrule
B elicitation gap & 0.850 & [0.814, 0.884] & 385 \\
A retrieval & 0.038 & [0.021, 0.055] & 17 \\
C integration & 0.044 & [0.027, 0.062] & 20 \\
A reasoning & 0.068 & [0.044, 0.095] & 31 \\
\bottomrule
\end{tabular}
\caption{Error decomposition of the 453 wrong Full-condition predictions on the categorical questions: elicitation gap (B), retrieval failure (A), integration failure (C), reasoning failure (A). Participant-clustered bootstrap CIs. Labels by Claude Opus 5.5 following the decision tree in Figure~\ref{fig:audits} (protocol in Appendix~\ref{app:e11-prompt}).}
\label{tab:e11}
\end{table}

\paragraph{Most errors are elicitation gaps.}
Table~\ref{tab:e11} assigns the \nWrongTraces{} wrong \full{} predictions on the categorical questions to the four leaves of Figure~\ref{fig:audits}. In \eBShare{} of them (95\% CI \eBCI) the auditor found no evidence in the transcript that discriminates the participant's option from the others. Retrieval failures (\eARetShare), integration failures (\eCIntShare) and reasoning failures (\eAReasShare) account for the rest; keeping only labels with confidence of at least 3 of 5 raises the elicitation-gap share to \eBShareHighConf. Within the elicitation-gap cases the auditor recorded related but non-discriminating passages, typically a general disposition that bears on the scenario without settling it, in \eBRelatedPct\% and nothing related at all in \eBNoMaterialPct\%. The predictor rarely overlooks or misreads evidence that is there; most of the time nothing in the transcript separates the options. The second annotator agreed with the first on \kCodexInformedAgree{} of the wrong traces ($\kappa$ = \kCodexInformedKappa{} over the four labels, \kCodexInformedKappaB{} for elicitation gap versus the rest) and put a larger share still in the elicitation gap (\kCodexInformedBShare). The \eAReasN{} reasoning failures are few but patterned: the predictor defaults to following rules, preserving harmony and honouring every obligation where participants chose pragmatically, disengaged or accepted a real trade-off, defaults with documented counterparts in aligned models \citep{cheung2025amplified,kim2026relational,dash2025motivated}; Appendix~\ref{app:biases} gives examples.\label{sec:results-biases}

\begin{table*}[tbp]
\centering
\footnotesize
\begin{tabular}{lcccc}
\toprule
Condition & Acc.\ (evidence) & Acc.\ (no evidence) & Difference [95\% CI] & $p$ \\
\midrule
Full & 0.431 (404) & 0.410 (378) & +0.021 [-0.047, +0.087] & 0.547 \\
Core-10 & 0.408 (404) & 0.394 (378) & +0.014 [-0.058, +0.086] & 0.701 \\
Summary & 0.406 (404) & 0.368 (378) & +0.038 [-0.040, +0.119] & 0.303 \\
Static (cohort 2) & 0.352 (213) & 0.428 (229) & -0.076 [-0.167, +0.015] & 0.115 \\
No-persona (draw mean) & 0.430 (404) & 0.401 (378) & +0.030 [-0.038, +0.095] & 0.363 \\
Majority (LOO) & 0.483 (404) & 0.370 (378) & +0.112 [+0.031, +0.199] & 0.002 \\
\bottomrule
\end{tabular}
\caption{Accuracy on the 17 categorical questions conditional on whether a blind audit (no access to answers or predictions) found decision-relevant evidence in the interviewee's transcript. Evidence available for 52\% of participant--question cells. Participant-clustered permutation tests; cluster-bootstrap CIs.}
\label{tab:e15}
\end{table*}

\paragraph{Evidence in the transcript does not make any predictor more accurate.}
An auditor who knows the outcome may find ``evidence'' in hindsight: the informed auditor found evidence for \eCtrlEvidencePct\% of \nControlTraces{} correct control predictions against \eWrongEvidencePct\% of the wrong ones. The blind audit removes this. Judging transcripts without access to answers or predictions, the auditor found decision-relevant evidence in \fCoveragePct\% of the \nBlindCells{} cells, most of it weak: only \fStrengthThreeN{} cells carry an explicit statement about the kind of situation the question describes. Table~\ref{tab:e15} crosses availability with correctness. For the \full{} regime, accuracy is \fFullAccYes{} when evidence exists and \fFullAccNo{} when it does not (difference \fFullDiff, 95\% CI \fFullCI, $p = \fFullP$); the picture is the same for \coreten, \summ{} and the model that never saw the transcript (\fNoPersonaDiff). The only predictor whose accuracy tracks evidence availability is the majority baseline (\fMajorityDiff, \fMajorityCI, $p = \fMajorityP$): the cells in which a transcript carries evidence are disproportionately cells in which the participant answered as most people did. Availability tracks conventionality more than it tracks the person. At the participant level, the share of a person's cells with blind evidence grows with transcript length (Spearman $\rho = \fRichnessTokensRho$, $p = \fRichnessTokensP$) but is unrelated to how accurately they are predicted (\full: $\rho = \fRichnessFullRho$, $p = \fRichnessFullP$).

\paragraph{The evidence that exists is not diagnostic.}
When evidence exists, the option it favours is the participant's answer in \fPrecisionPct\% of cells, above uniform guessing on these questions (\chanceChoiceMin{} to \chanceChoiceMax) but below the majority answer's hit rate on the same cells (\fMajorityAccYes): a reader who follows the transcript does worse than one who follows the crowd. The second annotator, judging the same cells blind, agreed on evidence availability in \kCodexBlindAgree{} of cells ($\kappa$ = \kCodexBlindKappa) and reached the same conclusion (coverage \kCodexBlindCoverage, precision \kCodexBlindPrecision; Appendix~\ref{app:audit-detail}). Together, the two audits locate the loss in a \emph{diagnosticity gap}: the interview elicits dispositions and values that bear on the scenarios, but seldom the situated policies that decide them.

\subsection{What the transcripts do carry}
\label{sec:results-traits}
\begin{table}[htbp]
\centering
\scriptsize
\setlength{\tabcolsep}{3pt}
\begin{tabular}{lccc}
\toprule
Target & Core-10 & Full & Full $-$ Core-10 [95\% CI] \\
\midrule
MBTI type (hit@2) & 0.326 & 0.543 & +0.217 [+0.043, +0.391] \\
Big Five match & 0.604 & 0.617 & +0.013 [-0.052, +0.078] \\
Decision accuracy & 0.395 & 0.400 & +0.005 [-0.017, +0.027] \\
\bottomrule
\end{tabular}
\caption{Within-participant effect of adding the adaptive follow-ups (Full $-$ Core-10; $N$=46) on MBTI type recovery (hit@2), binarised Big Five match (mean over five traits) and 25-question decision accuracy. Participant-bootstrap 95\% CIs.}
\label{tab:e16}
\end{table}

The transcripts are not uninformative. Table~\ref{tab:e16} compares, within participant, the effect of adding the adaptive follow-ups (\full{} $-$ \coreten) on three targets. MBTI type recovery (hit@2) rises from \mMbtiCore{} to \mMbtiFull{} (paired gain \mMbtiDiff, 95\% CI \mMbtiCI, $p = \mMbtiP$; \mMbtiUp{} participants improve and \mMbtiDown{} worsen), whereas binarised Big Five match moves by \mBigFiveDiff{} and decision accuracy by \mDecisionDiff{} (full personality metrics in Appendix~\ref{app:personality}). The follow-ups add type-level personality signal and no decision signal. MBTI self-reports are coarse and $N$ is small, so we treat the dissociation as supporting evidence; it agrees with the audits in that what an adaptive interview elicits is the material personality typologies summarise.

\section{Discussion}
\label{sec:discussion}

\paragraph{The diagnosticity gap.}
The interviews are not empty: a blind reader finds decision-related material in half of all cells, and they carry enough signal to raise MBTI type recovery. What they lack is material that discriminates between the options a scenario offers. People describe who they are in general terms (values, fears, coping styles, relational stances), and these generalities bear on most dilemmas without settling any of them. The situated policy that decides a particular trade-off is seldom articulated, and when it is, it is often a post-hoc rationalisation rather than a predictor of behaviour \citep{haidt2001emotional}. It explains why more of the same does not help: longer transcripts (Appendix~\ref{app:quantity}), adaptive follow-ups and summaries all add or compress general self-description, from which the predictor guesses no better than from nothing.

\paragraph{Decision-aware elicitation.}
If the bottleneck is diagnosticity, the design response is to elicit situated policies rather than richer self-portraits: brief scenario probes that mirror the structure of the target decision class (a forced trade-off under a stated constraint), asked about past behaviour rather than ideals, and sampled across the domains the downstream decisions span. Our fixed follow-ups moved in that direction and were cited more often than adaptive ones (\S\ref{sec:results-static}) without improving accuracy, so such probes must sit closer still to the target decisions, and their value should be tested as follow-ups were here: within subject, against the majority answer, with a pre-specified contrast.

\paragraph{Delegation.}
For practice the result is a caution rather than a verdict. It qualifies the headline of \citet{park2024generative}: an agent can reproduce much of a person's retest consistency on survey attitudes while saying little about which of several reasonable actions they would take. On value-laden decisions, a predictor conditioned on an hour of interview guesses what a particular person would do no better than one that knows nothing about them (Appendix~\ref{app:vignette} agrees). Persona agents built from such elicitation should not be delegated value-laden choices without checking back, and claims of individual fidelity should be read against the majority baseline.

\paragraph{Methodological lesson.}
Two easy measures of grounding are unreliable here: the predictor's self-reported evidence location does not measure what the transcript contains, and an informed auditor who knows the outcome finds ``evidence'' for correct predictions far more often than for wrong ones when a blind auditor finds no such difference. Studies of grounded or personalised prediction should treat self-reported grounding as a hypothesis, audit transcripts blind to outcomes, and prefer counterfactual tests of faithfulness \citep{wallat2024correctness,cohenwang2024contextcite} to citation checks.

\section{Conclusion}
\label{sec:conclusion}
A one-hour LLM-led interview, with adaptive or fixed follow-ups, did not make a model better at predicting a person's value-laden decisions than a model that knew nothing about the person, or than the majority answer. Two audits locate the loss in the transcripts: for most errors no option-discriminating evidence was found, and the related material that does exist matches the person's own answer less often than the majority answer does. The same transcripts improve MBTI type recovery. Interviews of this kind elicit traits, not decision policies; elicitation for individual decision simulation should be designed and tested for decisions.

\section*{Limitations}
\label{sec:limitations}
\paragraph{Sample.} \nParticipants{} participants is small for individual-level claims; the pre-specified contrasts have confidence intervals that exclude improvements larger than about three accuracy points, not smaller ones. Participants were recruited through campus networks and online media; other populations were not studied.

\paragraph{Audit labels.} The audits were performed by LLMs. Both annotators' protocols require verbatim quotations, which were validated automatically, and the two annotators agree substantially (Appendix~\ref{app:audit-detail}), but no human adjudicated either set and the second annotator shares a vendor with the predictor. The decision tree is new and its human reliability is unknown.

\paragraph{Models and confounds.} The main results use a single predictor (GPT-5 at medium reasoning effort). The grounding analysis was replicated with one open-weight predictor (Appendix~\ref{app:crossmodel}); the audits were not. Interviewer version (DeepSeek-R1 vs V4) is confounded with cohort and time, and the replication predictor shares a family with the cohort-2 interviewer.

\paragraph{Design of the control.} The fixed follow-up control has four questions at a somewhat smaller token budget than the adaptive follow-ups (median ratio \tokRatioStaticAdaptive), and it was answered after the decision questions, so order effects cannot be excluded; a budget-matched and order-balanced replication would strengthen the primary contrast. The interview--then--questions order shared by all regimes may also have primed participants' answers to the questions.

\paragraph{Task.} The \nItems{} questions are value-laden dilemmas of our own construction, each adapted from a published paradigm; other decision classes, such as risky choice under explicit probabilities or everyday consumption, are not represented, and the audits cover only the \nChoiceItems{} categorical questions. Three participants' core answers reached us rewritten by an LLM (one dictated them, one submitted a second-person restatement, one submitted summaries); excluding them changes no conclusion (Appendix~\ref{app:sensitivity}). MBTI ground truth is self-reported and often given as two types, so the personality dissociation is supporting evidence only.

\section*{Ethical Considerations}
\label{sec:ethics}
\paragraph{Review and consent.} The study involves voluntary adult participants, no deception, and no collection of protected attributes; it did not undergo formal IRB review, which the authors' institutions do not require for this design. All participants were informed of the purpose of the study, the scope of data collection, the research use of their de-identified responses, and their right to withdraw at any time. Participants in both cohorts received a small gift as compensation. Cohort-2 participants consented on the questionnaire to the public release of their de-identified transcripts and answers; cohort-1 participants were re-contacted for release consent. Participants who did not consent to release are excluded from the released data but not from the analyses.

\paragraph{Privacy.} Transcripts are stored under pseudonymous identifiers on encrypted local storage. Before release, transcripts are reviewed for identifying details (names, employers, places, dates), which are removed or generalised; the released audit labels quote only short spans. Demographic information was recorded only in aggregate at recruitment, cannot be linked to individual transcripts, and is reported in aggregate only (Appendix~\ref{app:demographics}); no demographic field is part of the release.

\paragraph{Risks.} Adaptive interviewing that elicits detailed personal narratives could be misused for profiling, and persona simulation deployed prematurely could produce confident but wrong inferences about individuals in consequential settings. Our results are themselves a caution against such deployment: interview-conditioned predictions of value-laden decisions were no better than baselines that know nothing about the person, and the predictor's default biases (Appendix~\ref{app:biases}) would systematically misrepresent people who deviate from majority-aligned patterns. The models studied are research subjects here, not a system we propose to deploy.

\bibliography{bib/custom,bib/new_refs}

\clearpage
\appendix
\renewcommand{\topfraction}{0.95}\renewcommand{\bottomfraction}{0.95}\renewcommand{\textfraction}{0.05}
\renewcommand{\floatpagefraction}{0.85}\renewcommand{\dbltopfraction}{0.95}\renewcommand{\dblfloatpagefraction}{0.85}
\setcounter{topnumber}{4}\setcounter{bottomnumber}{3}\setcounter{totalnumber}{6}\setcounter{dbltopnumber}{3}
\FloatBarrier
\section{Prompts}
\label{app:prompts}

\subsection{Interview meta-prompt (verbatim)}
\label{app:meta-prompt}
The interviewer was driven by the following meta-prompt.
\begin{Verbatim}[breaklines=true,breakanywhere=true,fontsize=\footnotesize]
You are a deep persona interviewer whose purpose is to understand the participant
as completely as possible — their reasoning patterns, fears, values, blind spots,
and worldview.

Your interview proceeds in three stages:
---
Stage 1 — Generate Ten Core Questions
1. Generate exactly 10 open-ended, concrete, introspective questions that—taken
   together—let you form a 360-degree picture of the participant.
2. Each question must correspond to one of these ten thematic domains (use each once):
   1. Behavioral Priorities & Trade-offs
   2. Decision Logic & Value Hierarchy
   3. Fears & Deep Motivation
   4. Self-Awareness & Blind Spots
   5. Stress & Coping Mechanisms
   6. Relational Logic & Boundaries
   7. Value Conflicts & Compromise
   8. Core Identity & Self-Definition
   9. Emotional Vulnerability & Regulation
   10. Life Narrative & Sense of Meaning
3. Questions must be specific and story-eliciting, avoiding abstract or yes/no phrasing.
4. Tone: natural, empathic, intellectually curious.

Output template:
Stage 1 – Ten Questions:
1. ...
...
10. ...
---
Stage 2 — Adaptive Follow-Up (trigger after participant answers)
After the participant provides answers to all ten questions, you will:
1. Read all ten answers carefully.
2. Identify 3–5 aspects that seem unclear, internally inconsistent, emotionally charged,
   or highly characteristic.
3. For each, craft a targeted follow-up question to deepen or clarify your understanding.
4. Reference the participant's words when possible.

Output template (after answers received):
Stage 2 – Follow-Up Questions:
F1. [follow-up + brief note of purpose]
F2. ...
F3. ...
---
Interaction style
- Do not give analysis or labels before Stage 2.
- Remain open, gentle, and precise.
- When Stage 2 follow-ups are answered, you may summarize the participant's personality
  insights per domain (stage3).
\end{Verbatim}

\paragraph{Psychological basis of the ten domains.}
\begin{table}[h]
\centering\footnotesize
\setlength{\tabcolsep}{3pt}\begin{tabular}{cp{2.5cm}p{3.6cm}}
\toprule
\# & Domain & Basis \\
\midrule
1 & Behavioural priorities \& trade-offs & value enactment \citep{schwartz1992universals} \\
2 & Decision logic \& value hierarchy & moral reasoning, cognitive style \citep{rest1986moral,baron1994thinking} \\
3 & Fears \& deep motivation & reinforcement sensitivity \citep{gray1987psychology} \\
4 & Self-awareness \& blind spots & self--other knowledge asymmetry \citep{vazire2010self} \\
5 & Stress \& coping & trait--coping meta-analysis \citep{connor2007psychological} \\
6 & Relational logic \& boundaries & attachment, interpersonal theory \citep{bowlby1988secure,bartholomew1991attachment} \\
7 & Value conflicts \& compromise & integrative complexity \citep{tetlock1986reasoning} \\
8 & Core identity & narrative identity \citep{mcadams2006identity} \\
9 & Emotional vulnerability \& regulation & emotion regulation \citep{gross2003emotion} \\
10 & Life narrative \& meaning & meaning-making \citep{mclean2007meaning} \\
\bottomrule
\end{tabular}
\caption{The ten interview domains fixed in the meta-prompt and their psychological basis.}
\label{tab:persona_domains}
\end{table}

\subsection{Fixed follow-up questions (Static regime)}
\label{app:static}
\begin{table}[htbp]
\centering
\small
\setlength{\tabcolsep}{3pt}
\begin{tabular}{lp{0.6\linewidth}c}
\toprule
 & Fixed follow-up question & Valid answers \\
\midrule
S1 & Describe a time when real-world conditions (such as money, time, responsibilities or other people's expectations) forced you to compromise on something you cared about. How did you make the trade-off in the end? & 25 \\
S2 & Tell me about a time you had to choose between what you wanted and what someone close to you needed. What did you do, and why? & 24 \\
S3 & Name a principle or rule you usually follow, and tell me about a time you broke it. What happened? & 23 \\
S4 & When you strongly disagree with a friend or colleague, what usually happens next? Give a concrete example. & 26 \\
\bottomrule
\end{tabular}
\caption{The four fixed follow-up questions answered by every cohort-2 participant after the dilemma questions (the Static regime). Participants were asked for at least 40 characters per answer; non-informative answers (for example that no such experience came to mind) were dropped for that question.}
\label{tab:static}
\end{table}

\subsection{Prediction prompts}
\label{app:pred-prompt}
The \coreten, \full{} and \static{} regimes use two calls per participant and regime.

\paragraph{Call 1, prediction.}
\begin{Verbatim}[breaklines=true,breakanywhere=true,fontsize=\footnotesize]
[System]
You are simulating the decision-making of a specific interviewee. You must use only the
persona information provided; do not bring in outside common sense or additional
assumptions. If the persona information is insufficient to support a conclusion, do not
make one up. When asked for evidence quotations, copy short phrases verbatim from the
persona information; if none can be found, output None.

[User]
Below is information collected from an interview with one interviewee.

=== Persona Information ===
{conversation_content}

Notes:
- The persona information may contain both Core Interview and Follow-up content.
- If the text explicitly distinguishes "Core Interview" from "Follow-up", judge the
  source of each quotation by that distinction.
- If it does not, treat all persona information as Core Interview.

Now answer all of the decision questions below as this interviewee.

=== Decision Questions ===
{question_content}

Output strictly in the following format (no other headings or fields):

### 1) Brief reasoning (2-4 sentences)
Explain in 2-4 sentences why this interviewee would choose as they do. Use only the
persona information above; unless it offers no clue at all, avoid generic moral or
common-sense statements.

### 2) Chosen option
Follow each question's own options and format: option letter for choice questions, an
integer within the scale for rating questions, a complete ordering for the ranking
question.

### 3) Evidence quotations and source
Give at most two verbatim quotations from the persona information (each at most 25 words)
that directly support the reasoning. For each: - Quotation: "..."
If no quotation in the persona information directly supports the reasoning, output None.

### 4) Confidence
Choose exactly one:
- certain: clear, direct clues support the choice; almost no equivalent alternative
  interpretation exists.
- not fully certain: some clues, but other options or interpretations cannot be ruled out.
- cannot determine: almost no relevant clues; the answer relies mainly on generalisation
  or common-sense completion.
\end{Verbatim}

\paragraph{Call 2, annotation.}
\begin{Verbatim}[breaklines=true,breakanywhere=true,fontsize=\footnotesize]
[System]
You are an annotation assistant. Your task is to label a given model output. Do not change
the decision, add evidence, or rewrite the reasoning. Output only the requested labels, in
the required format.

[User]
You receive: (1) the persona information, (2) the decision questions, (3) the output of
Prompt 1 (reasoning / option / evidence quotations, in question order).

=== Persona Information ===
{conversation_content}
=== Decision Questions ===
{question_content}
=== Prompt 1 output ===
{model_output}

Label A: reasoning category (one dominant category per question)
- narrative: reasons mainly from specific personal experiences, stories or past events
- value: reasons mainly from abstract preferences, principles, values or what matters more
- constraints: reasons mainly from rules, obligations, responsibilities, limits, practical
  conditions or feasibility
- generic: hardly based on the persona information; common-sense, vague or sweeping
  statements

Label B: evidence location (one per question), decided from the quotations listed in
Prompt 1 and their declared sources
- only Core Interview quotations -> Core Interview
- only Follow-up quotations -> Follow-up
- at least one of each -> Both
- evidence None -> None

Output strictly in the JSON format provided, with no extra fields and no explanatory text.
\end{Verbatim}

\subsection{Summary-regime prompt}
\label{app:summary-prompt}
\begin{Verbatim}[breaklines=true,breakanywhere=true,fontsize=\footnotesize]
[System]
You are predicting how a specific interviewee would answer the questions below. The only
background about this person is the personality-insight summary provided, generated by the
interviewer model. Base your predictions on the value orientations, preferences,
behavioural patterns and coping styles described in the summary. Treat the summary as a
general description of the interviewee; do not add interview content or personal
experiences the summary does not provide, and do not substitute your own preferences for
the interviewee's. Give one prediction per question in the required format and explain in
2-4 sentences how it relates to the characteristics described. If the summary offers
limited cues, still give the most likely prediction and say that the basis is limited; do
not invent background. Return only the specified JSON: answer and reasoning per question,
no evidence quotations, locations or confidence labels.

[User]
The personality-insight summary below is the only background for predicting this
interviewee's answers.
=== Personality Summary ===
{personality_summary}
=== Decision Questions ===
{question_content}
Predict this interviewee's answers to all 25 questions (1-25). The answer field must follow
each question's own format: option letter; integer within the scale; a complete ordering
of A, B, C, D, E with no separators. The reasoning field explains in 2-4 sentences which
characteristics in the summary make the prediction plausible; say so when the basis is
limited. Return one JSON object whose only top-level key is "answers", mapping "1" to "25"
to objects with answer and reasoning only.
\end{Verbatim}
The personality version asks instead for the two most likely MBTI types in order and a 0--40 score per Big Five trait.

\subsection{Zero-context prompt}
\label{app:nopersona-prompt}
\begin{Verbatim}[breaklines=true,breakanywhere=true,fontsize=\footnotesize]
Please answer the following scenario questions as an ordinary respondent. Treat the
described situation as your own and choose what you would most likely do. Do not add
personal background or experiences the question does not provide. Follow each question's
options and answer format, and explain your choice briefly in 2-4 sentences.

{question_content}
\end{Verbatim}

\subsection{Informed audit protocol}
\label{app:e11-prompt}
\begin{Verbatim}[breaklines=true,breakanywhere=true,fontsize=\footnotesize]
You are auditing predictions made by a language model that was asked to predict how a
specific interviewee would answer multiple-choice dilemma questions, using only that
interviewee's interview transcript (ten core questions with answers, then adaptive
follow-up questions with answers). For each trace you receive: the question with its
options, the interviewee's TRUE answer, the model's PREDICTED answer, the model's
reasoning, the evidence excerpt the model cited, and the model's own label of where that
evidence came from. The full transcript is provided once per interviewee. Quote it verbatim.

For traces where the prediction was WRONG: answer in order, stop at the first terminal label
1. Evidence existence. Does the transcript contain decision-relevant evidence for this
   question — a passage that would materially shift a careful human reader's probability
   of at least one option (not necessarily the true one)?
   - If the evidence supports the predicted (wrong) answer as strongly as the true answer,
     the transcript does not discriminate -> treat as NO.
   - If the interviewee's true answer contradicts the values/constraints they themselves
     state in the transcript, the answer was not predictable from elicited evidence -> NO,
     and add the flag inconsistent_participant.
   - NO -> label B-elicitation-gap.
2. Retrieval. Does the model's cited excerpt include that evidence (verbatim or clear
   semantic overlap)? An empty excerpt or an excerpt about something else -> NO ->
   label A-retrieval.
3. Error form.
   - The model relied on a single fragment while two or more segments from DIFFERENT
     interview turns jointly indicate a value or constraint it missed -> C-integration.
   - The cited evidence is apt but the inference drawn from it is wrong -> A-reasoning.

For CONTROL traces (the prediction was CORRECT): answer only step 1 and report the spans:
label CONTROL-evidence if decision-relevant evidence exists, CONTROL-no-evidence otherwise.

Rules: quote evidence verbatim as substrings of the transcript with block and turn; never
paraphrase inside a quote; do not infer facts that are not in the transcript; do not use
the true answer as evidence of what the transcript says; confidence is 1-5; rationale
<= 60 words. Output one JSON object per trace.
\end{Verbatim}

\subsection{Blind audit protocol}
\label{app:e15-prompt}
\begin{Verbatim}[breaklines=true,breakanywhere=true,fontsize=\footnotesize]
You are judging what an interview transcript can tell a careful reader about how the
interviewee would answer multiple-choice dilemma questions. You see the interviewee's full
transcript (ten core questions with answers, then follow-up questions with answers)
and 17 questions with their options. You do NOT see the interviewee's actual
answers and you do NOT see any model's predictions — judge the transcript alone.

For EACH question:
1. evidence_exists — does the transcript contain decision-relevant evidence: a passage
   that would materially shift a careful reader's probability for at least one option?
   Passages that support all options about equally, or that are only about the general
   topic without bearing on this specific trade-off, do NOT count.
2. strength — 0 none; 1 indirect (a general disposition that bears on the trade-off);
   2 moderate (a stated value or habit that clearly applies); 3 explicit (the interviewee
   describes this kind of situation or states a rule for it).
3. favored_options — the option letter(s) the evidence favours (one or more; empty when
   evidence_exists is false). If, forced to answer from the transcript alone, you would
   pick one option, put it FIRST.
4. spans — verbatim quotes (substrings of the transcript; block core or followup; turn).
5. note — <= 40 words on how the evidence bears on the question.

Rules: quote verbatim; do not infer facts not in the transcript; do not guess what "most
people" would do — only what THIS transcript says; judge every one of the 17 questions.
\end{Verbatim}

\FloatBarrier
\section{Decision questions}
\label{app:items}
\begin{table*}[tbp]
\centering
\footnotesize
\setlength{\tabcolsep}{3pt}
\begin{tabular}{llp{2.6cm}p{6.2cm}cc}
\toprule
Question & Type & Domain & Source paradigm(s) & Options & Chance \\
\midrule
Q1 & Choice & Decision making & Delay Discounting, CNI trolley (omission/commission weighting) & 4 & 0.25 \\
Q2 & Choice & Emotion regulation & ERQ (response selection framing) & 4 & 0.25 \\
Q3 & Choice & Decision making & Delay Discounting (MCQ-style) & 4 & 0.25 \\
Q4 & Choice & Moral reasoning & Third-Party Punishment, CNI trolley & 4 & 0.25 \\
Q5 & Likert & Emotion regulation & ERQ & 5 & 0.20 \\
Q6 & Choice & Value & Schwartz Values (benevolence vs self-direction), ERQ (self-regulation plan) & 4 & 0.25 \\
Q7 & Likert & Value & SCC & 5 & 0.20 \\
Q8 & Choice & Social cooperation \& fairness & Dictator Game/Public Goods & 4 & 0.25 \\
Q9 & Choice & Social cooperation \& fairness & Dictator/Public Goods & 4 & 0.25 \\
Q10 & Choice & Moral reasoning & CNI trolley, (omission/commission \& aggregation) & 4 & 0.25 \\
Q11 & Likert & Value & Schwartz Values (security vs achievement) & 5 & 0.20 \\
Q12 & Likert & Value & Schwartz Values (openness vs security) & 5 & 0.20 \\
Q13 & Choice & Social cooperation \& fairness & Ultimatum Game & 4 & 0.25 \\
Q14 & Choice & Decision making & Delay Discounting & 3 & 0.33 \\
Q15 & Choice & Decision making & Delay Discounting & 3 & 0.33 \\
Q16 & Likert & Emotion regulation & ERQ & 5 & 0.20 \\
Q17 & Ranking & Value & Schwartz Values & -- & 0.50 \\
Q18 & Likert & Value & SCC, Schwartz Values & 5 & 0.20 \\
Q19 & Choice & Moral reasoning & Public Goods, CNI trolley (aggregation) & 4 & 0.25 \\
Q20 & Likert & Value & MLQ, Schwartz Values & 5 & 0.20 \\
Q21 & Choice & Moral reasoning & CNI trolley, Procedural Justice vs Substantive Justice & 4 & 0.25 \\
Q22 & Choice & Social cooperation \& fairness & Third-Party Fairness, Social Inclusion Dilemmas & 4 & 0.25 \\
Q23 & Choice & Personality trait & Conflict Management Style, Big Five Agreeableness & 4 & 0.25 \\
Q24 & Choice & Personality trait & Role Conflict Paradigm, Big Five Conscientiousness & 4 & 0.25 \\
Q25 & Choice & Moral reasoning & Moral Personality Conflict, Big Five Trait Interaction & 4 & 0.25 \\
\bottomrule
\end{tabular}
\caption{The 25 dilemma questions: response type, thematic domain (Table~\ref{tab:domains}), the psychological paradigm(s) each question was adapted from, number of options and chance accuracy under uniform guessing. Full question texts are released with the data.}
\label{tab:items}
\end{table*}

\begin{table*}[tbp]
\centering
\footnotesize
\setlength{\tabcolsep}{4pt}
\begin{tabular}{lp{6cm}p{6cm}}
\toprule
Pair & First question: option $\rightarrow$ stance & Second question: option $\rightarrow$ stance \\
\midrule
Q8/Q9 & A: equal; B, C: by contribution; D: keep & A: by contribution; B: charity; C: equal; D: keep \\
Q11/Q12 & 4--5: privacy first; 1--2: utility first; 3: neutral & 4--5: utility first; 1--2: privacy first; 3: neutral \\
Q14/Q15 & A: smaller now; B: larger later; C: both & A: larger later; B: smaller now; C: both \\
Q19/Q20 & A: local; B: global; C: split; D: no donation & 4--5: local; 1--2: global; 3: split \\
Q21/Q25 & A: rule; B: outcome; C: lawful middle path; D: defer & A: rule; B: outcome; C: lawful middle path; D: defer \\
\bottomrule
\end{tabular}
\caption{Stance mappings used to score the consistency probes. A participant or predictor is consistent on a pair when the two answers map to the same stance; options without a counterpart (charity in Q9, no donation in Q19) count as inconsistent.}
\label{tab:e18_mapping}
\end{table*}

\begin{table}[htbp]
\centering
\footnotesize
\setlength{\tabcolsep}{3pt}
\begin{tabular}{lp{3.3cm}}
\toprule
Domain & Questions \\
\midrule
Moral reasoning & 4, 10, 19, 21, 25 \\
Social cooperation \& fairness & 8, 9, 13, 22 \\
Emotion regulation & 2, 5, 16 \\
Decision making & 1, 3, 14, 15 \\
Value & 6, 7, 11, 12, 17, 18, 20 \\
Personality trait & 23, 24 \\
\bottomrule
\end{tabular}
\caption{Thematic domains of the 25 dilemma questions.}
\label{tab:domains}
\end{table}

The question set also contains five mirrored consistency-probe pairs, (Q8, Q9), (Q11, Q12), (Q14, Q15), (Q19, Q20) and (Q21, Q25), which pose the same trade-off in two guises with permuted options, a reverse-coded statement, a reversed reward domain or a choice question mirrored by a rating. Each answer is mapped to a stance (Table~\ref{tab:e18_mapping}), and a participant or predictor is consistent on a pair when both answers map to the same stance. Participants themselves are consistent on only \qHumanConsistent{} of pairs. Table~\ref{tab:e18} shows that every interview regime predicts more consistently than the participants behave (\qConsistentFull{} for \full), that whether a given participant is consistent is matched at about chance (agreement \qAgreeFull{} for \full), and that no regime reproduces a participant's pair of stances much more often than the baselines do.
\begin{table}[htbp]
\centering
\scriptsize
\setlength{\tabcolsep}{3pt}
\begin{tabular}{lcccc}
\toprule
Regime & $N$ & Consistent & Agreement & Reproduced \\
\midrule
\multicolumn{5}{l}{\textit{Cohort 1 (participants: 0.33)}} \\
Core-10 & 20 & 0.550 & 0.400 & 0.130 \\
Full & 20 & 0.470 & 0.520 & 0.190 \\
Summary & 20 & 0.550 & 0.500 & 0.180 \\
No-persona & 20 & 0.600 & 0.410 & 0.200 \\
Majority & 20 & 0.250 & 0.520 & 0.090 \\
\multicolumn{5}{l}{\textit{Cohort 2 (participants: 0.32)}} \\
Core-10 & 26 & 0.500 & 0.515 & 0.215 \\
Full & 26 & 0.492 & 0.508 & 0.185 \\
Static & 26 & 0.546 & 0.485 & 0.192 \\
Summary & 26 & 0.538 & 0.462 & 0.185 \\
No-persona & 26 & 0.600 & 0.508 & 0.169 \\
Majority & 26 & 0.246 & 0.569 & 0.154 \\
\multicolumn{5}{l}{\textit{Pooled (participants: 0.33)}} \\
Core-10 & 46 & 0.522 & 0.465 & 0.178 \\
Full & 46 & 0.483 & 0.513 & 0.187 \\
Static & 26 & 0.546 & 0.485 & 0.192 \\
Summary & 46 & 0.543 & 0.478 & 0.183 \\
No-persona & 46 & 0.600 & 0.465 & 0.183 \\
Majority & 46 & 0.248 & 0.548 & 0.126 \\
\bottomrule
\end{tabular}
\caption{Consistency probes (five mirrored pairs; stance mappings in Table~\ref{tab:e18_mapping}). Pred.\ consistent: share of participants for whom the predictions on the two questions map to the same stance; agreement: share for whom the prediction's consistent/inconsistent status matches the participant's own; pair reproduced: share for whom both predicted stances equal the participant's. Means over the five pairs.}
\label{tab:e18}
\end{table}

\FloatBarrier
\section{Additional results}
\label{app:results}

\subsection{Accuracy by cohort and by question type}
\label{app:types}
\begin{table}[htbp]
\centering
\footnotesize
\setlength{\tabcolsep}{3pt}
\begin{tabular}{lcccc}
\toprule
Condition & $N$ & Acc & 95\% CI & $\Delta$ \\
\midrule
\textit{Cohort 1} &  &  &  &  \\
Majority (LOO) & 20 & 0.358 & [0.324, 0.395] & -0.069 \\
No-persona & 20 & 0.427 & [0.390, 0.463] & -- \\
Core-10 & 20 & 0.401 & [0.359, 0.444] & -0.027 \\
Full & 20 & 0.399 & [0.351, 0.450] & -0.029 \\
Summary & 20 & 0.393 & [0.350, 0.433] & -0.034 \\
Static & -- & -- & -- & -- \\
\textit{Cohort 2} &  &  &  &  \\
Majority (LOO) & 26 & 0.426 & [0.385, 0.466] & +0.061 \\
No-persona & 26 & 0.365 & [0.309, 0.413] & -- \\
Core-10 & 26 & 0.391 & [0.345, 0.434] & +0.026 \\
Full & 26 & 0.402 & [0.361, 0.435] & +0.037 \\
Summary & 26 & 0.385 & [0.336, 0.432] & +0.020 \\
Static & 26 & 0.394 & [0.348, 0.439] & +0.030 \\
\bottomrule
\end{tabular}
\caption{Overall accuracy by cohort (same structure as Table~\ref{tab:table1}; $\Delta$ = difference from the cohort's No-persona accuracy). No-persona rows are GPT-5 (medium) zero-context draws in each cohort's own question wording.}
\label{tab:table1_cohort}
\end{table}

\begin{table}[htbp]
\centering
\small
\setlength{\tabcolsep}{3pt}
\begin{tabular}{lcc}
\toprule
Condition & Exact & Off-by-1 \\
\midrule
Pooled: No-persona & 0.286 & 0.814 \\
Pooled: Core-10 & 0.326 & 0.714 \\
Pooled: Full & 0.301 & 0.677 \\
Pooled: Summary & 0.339 & 0.742 \\
Cohort 1: No-persona & 0.314 & 0.821 \\
Cohort 1: Core-10 & 0.321 & 0.707 \\
Cohort 1: Full & 0.293 & 0.679 \\
Cohort 1: Summary & 0.329 & 0.721 \\
Cohort 2: No-persona & 0.264 & 0.808 \\
Cohort 2: Core-10 & 0.330 & 0.720 \\
Cohort 2: Full & 0.308 & 0.676 \\
Cohort 2: Summary & 0.346 & 0.758 \\
Cohort 2: Static & 0.341 & 0.731 \\
\bottomrule
\end{tabular}
\caption{Likert questions (7): exact-match accuracy and off-by-one tolerance.}
\label{tab:c3}
\end{table}

\begin{table}[htbp]
\centering
\small
\setlength{\tabcolsep}{3pt}
\begin{tabular}{lcccc}
\toprule
Question & No-persona & Core-10 & Full & Summary \\
\midrule
Q1 & 0.326 & 0.326 & 0.326 & 0.283 \\
Q2 & 0.391 & 0.391 & 0.435 & 0.435 \\
Q3 & 0.565 & 0.478 & 0.478 & 0.370 \\
Q4 & 0.174 & 0.283 & 0.283 & 0.261 \\
Q5 & 0.370 & 0.326 & 0.391 & 0.413 \\
Q6 & 0.652 & 0.391 & 0.565 & 0.457 \\
Q7 & 0.326 & 0.370 & 0.326 & 0.283 \\
Q8 & 0.348 & 0.413 & 0.370 & 0.370 \\
Q9 & 0.370 & 0.500 & 0.457 & 0.391 \\
Q10 & 0.220 & 0.217 & 0.239 & 0.261 \\
Q11 & 0.283 & 0.304 & 0.174 & 0.348 \\
Q12 & 0.239 & 0.239 & 0.196 & 0.283 \\
Q13 & 0.248 & 0.304 & 0.261 & 0.196 \\
Q14 & 0.380 & 0.391 & 0.435 & 0.457 \\
Q15 & 0.374 & 0.370 & 0.435 & 0.304 \\
Q16 & 0.326 & 0.348 & 0.326 & 0.391 \\
Q17 & 0.731 & 0.776 & 0.750 & 0.754 \\
Q18 & 0.239 & 0.283 & 0.370 & 0.304 \\
Q19 & 0.326 & 0.174 & 0.196 & 0.152 \\
Q20 & 0.217 & 0.413 & 0.326 & 0.348 \\
Q21 & 0.370 & 0.348 & 0.391 & 0.391 \\
Q22 & 0.478 & 0.500 & 0.500 & 0.500 \\
Q23 & 0.609 & 0.543 & 0.500 & 0.500 \\
Q24 & 0.652 & 0.630 & 0.696 & 0.674 \\
Q25 & 0.587 & 0.565 & 0.587 & 0.587 \\
\bottomrule
\end{tabular}
\caption{Per-question accuracy by condition (pooled).}
\label{tab:c4}
\end{table}

\begin{figure*}[t]
\centering
\includegraphics[width=\textwidth]{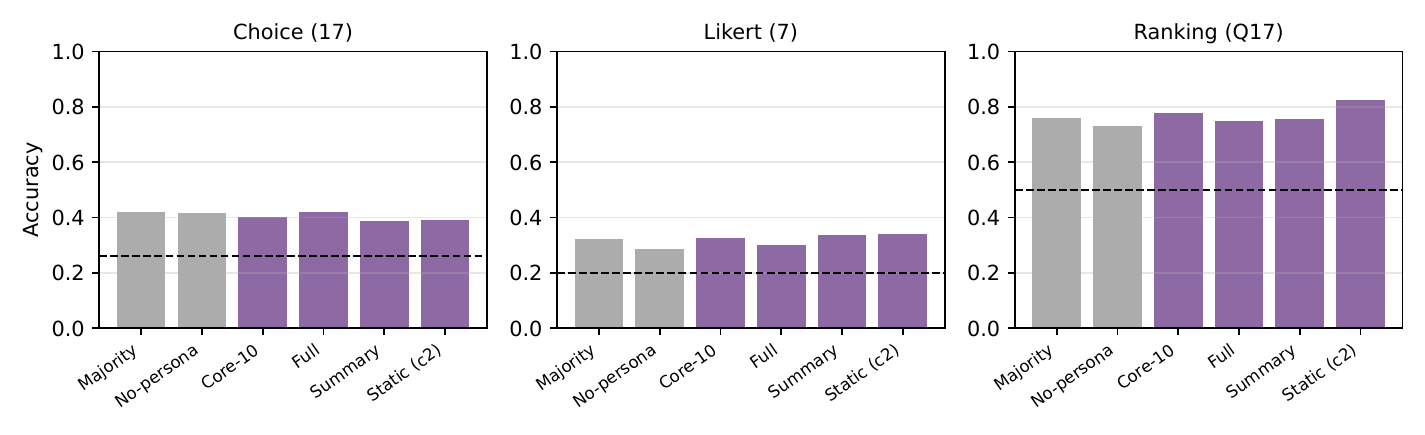}
\caption{Accuracy by question type for each input regime and baseline, both cohorts pooled (Static: cohort 2 only). Dashed lines: expected accuracy under uniform guessing for that question type.}
\label{fig:qtypes}
\end{figure*}

Table~\ref{tab:table1_cohort} gives Table~\ref{tab:table1} per cohort; the zero-context rows use GPT-5 draws in each cohort's own question wording. An earlier set of draws from GPT-5.1 at the same reasoning effort scores \accNoPersonaAltGptFivePointOneCohortOne{} on cohort 1 and \accNoPersonaAltGptFivePointOneCohortTwo{} on cohort 2, so the model version does not change the baseline. Table~\ref{tab:c3} reports Likert accuracy with a one-point tolerance and Table~\ref{tab:c4} per-question accuracy. Figure~\ref{fig:qtypes} breaks accuracy down by question type: ranking (Q17, scored by pairwise order agreement) is high for every regime and baseline alike, choice questions sit near the majority answer, and Likert questions are close to chance under exact match.

\subsection{Pre-specified contrasts by cohort and evidence use}
\label{app:contrasts}
\begin{table*}[tbp]
\centering
\footnotesize
\setlength{\tabcolsep}{3pt}
\begin{tabular}{lccccccc}
\toprule
Contrast & $N$ & Acc A & Acc B & A$-$B [95\% CI] & $p$ & $p_{\mathrm{Holm}}$ & Effect \\
\midrule
Core-10 vs No-persona (cohort 1) & 20 & 0.401 & 0.427 & -0.027 [-0.071, +0.015] & 0.246 & -- & $d$=-0.27, $h$=-0.05 \\
Core-10 vs No-persona (cohort 2) & 26 & 0.391 & 0.365 & +0.026 [-0.020, +0.072] & 0.283 & -- & $d$=0.21, $h$=0.05 \\
Full vs Core-10 (cohort 1) & 20 & 0.399 & 0.401 & -0.002 [-0.039, +0.036] & 0.918 & -- & $d$=-0.02, $h$=-0.00 \\
Full vs Core-10 (cohort 2) & 26 & 0.402 & 0.391 & +0.011 [-0.016, +0.035] & 0.441 & -- & $d$=0.16, $h$=0.02 \\
Static vs Core-10 (cohort 2) & 26 & 0.394 & 0.391 & +0.003 [-0.022, +0.029] & 0.824 & -- & $d$=0.05, $h$=0.01 \\
Full vs No-persona (cohort 2) & 26 & 0.402 & 0.365 & +0.037 [-0.005, +0.079] & 0.098 & -- & $d$=0.34, $h$=0.08 \\
Full vs No-persona (pooled) & 46 & 0.400 & 0.392 & +0.008 [-0.024, +0.041] & 0.622 & -- & $d$=0.07, $h$=0.02 \\
Full vs Summary (pooled) & 46 & 0.400 & 0.388 & +0.012 [-0.017, +0.040] & 0.422 & -- & $d$=0.12, $h$=0.02 \\
Grounded vs core-only (Full, cat., cohort 1) & 20 & 0.455 & 0.393 & +0.062 [-0.065, +0.196] & 0.324 & -- & $h$=0.13 \\
Grounded vs core-only (Full, cat., cohort 2) & 26 & 0.437 & 0.401 & +0.036 [-0.072, +0.144] & 0.475 & -- & $h$=0.07 \\
Static-grounded vs core-only (Static, cat., c2) & 26 & 0.389 & 0.372 & +0.018 [-0.062, +0.093] & 0.811 & -- & $h$=0.04 \\
\bottomrule
\end{tabular}
\caption{Per-cohort and exploratory versions of the contrasts in Table~\ref{tab:e7} (not Holm-corrected).}
\label{tab:e7_appendix}
\end{table*}

\begin{table*}[tbp]
\centering
\scriptsize
\setlength{\tabcolsep}{3pt}
\begin{tabular}{lcccc}
\toprule
Condition & Follow-up used & Grounded acc.\ ($n$) & Core-only acc.\ ($n$) & Transitions: $+$ / $-$ / $=$right / $=$wrong \\
\midrule
Both cohorts: Full, adaptive follow-ups & 0.385 & 0.445 (301) & 0.397 (443) & 32 / 21 / 102 / 146 \\
Cohort 1: Full, adaptive follow-ups & 0.394 & 0.455 (134) & 0.393 (201) & 15 / 6 / 46 / 67 \\
Cohort 2: Full, adaptive follow-ups & 0.378 & 0.437 (167) & 0.401 (242) & 17 / 15 / 56 / 79 \\
Cohort 2: Static, fixed follow-ups & 0.593 & 0.389 (262) & 0.372 (156) & 22 / 21 / 80 / 139 \\
\bottomrule
\end{tabular}
\caption{Evidence use on the 17 categorical questions: share of traces citing follow-up (adaptive or static) evidence, accuracy of follow-up-grounded vs core-only traces, and paired transitions against the Core-10 prediction for the same participant--question. Traces with no evidence location are excluded from the grounded/core-only split.}
\label{tab:d}
\end{table*}

\begin{table*}[tbp]
\centering
\footnotesize
\setlength{\tabcolsep}{4pt}
\begin{tabular}{lccc}
\toprule
Full $-$ Core-10 (paired) & MBTI hit@2 & Big Five match & Decision accuracy \\
\midrule
cohort 1 ($N$=20) & +0.200 [-0.100, +0.500] & +0.020 [-0.110, +0.140] & -0.002 [-0.040, +0.037] \\
cohort 2 ($N$=26) & +0.231 [+0.038, +0.423] & +0.008 [-0.054, +0.069] & +0.011 [-0.016, +0.035] \\
pooled ($N$=46) & +0.217 [+0.043, +0.391] & +0.013 [-0.052, +0.078] & +0.005 [-0.017, +0.027] \\
\bottomrule
\end{tabular}
\caption{Table~\ref{tab:e16} by cohort: within-participant effect of adding adaptive follow-ups on MBTI type recovery (hit@2), binarised Big Five match and 25-question decision accuracy. Participant-bootstrap 95\% CIs.}
\label{tab:e16_cohort}
\end{table*}

\begin{table}[htbp]
\centering
\scriptsize
\setlength{\tabcolsep}{3pt}
\begin{tabular}{lccc}
\toprule
Predictor & Grounded & Core-only & Difference [95\% CI] \\
\midrule
Full & 0.445 & 0.397 & +0.048 [-0.034, +0.131] \\
Core-10 & 0.409 & 0.388 & +0.020 [-0.065, +0.106] \\
No-persona & 0.433 & 0.406 & +0.027 [-0.055, +0.106] \\
Majority & 0.475 & 0.402 & +0.073 [+0.001, +0.151] \\
\bottomrule
\end{tabular}
\caption{Accuracy on the categorical cells where the Full-condition trace cites follow-up evidence (grounded, $n$=301) versus cells where it cites only core evidence (core-only, $n$=443), computed for Full itself and for predictors that never saw the follow-ups. Participant-cluster bootstrap CIs.}
\label{tab:e17}
\end{table}

Table~\ref{tab:e17} measures the follow-up-grounded versus core-only gap on the same cells for predictors that never saw the follow-ups; the majority answer shows the largest gap, so the cells on which the predictor cites follow-ups are simply easier. On the cells where the \full{} trace cites follow-ups, \full{} reaches \jFullGrounded{} against \jCoreGrounded{} for the \coreten{} prediction of the same participant on the same question (the paired transitions are in \S\ref{sec:results-static}).
In the primary contrast, adaptive follow-ups gave the higher accuracy for \cFullStaticPartA{} of the \nCohortTwo{} participants and fixed follow-ups for \cFullStaticPartB. Table~\ref{tab:d} gives evidence use and the paired transitions against the \coreten{} prediction per cohort.

\subsection{Cross-model replication of the grounding analysis}
\label{app:crossmodel}
\begin{table}[htbp]
\centering
\scriptsize
\setlength{\tabcolsep}{3pt}
\begin{tabular}{lccc}
\toprule
Predictor & Grounded & Core-only & Difference [95\% CI] \\
\midrule
DeepSeek Full & 0.379 & 0.347 & +0.032 [-0.046, +0.106] \\
DeepSeek Core-10 & 0.336 & 0.378 & -0.042 [-0.121, +0.036] \\
GPT-5 Full & 0.412 & 0.430 & -0.018 [-0.081, +0.042] \\
No-persona (DeepSeek) & 0.398 & 0.387 & +0.010 [-0.046, +0.062] \\
GPT-5 Core-10 & 0.394 & 0.409 & -0.015 [-0.084, +0.053] \\
No-persona (GPT-5) & 0.423 & 0.408 & +0.015 [-0.054, +0.080] \\
Majority & 0.419 & 0.422 & -0.003 [-0.075, +0.068] \\
\bottomrule
\end{tabular}
\caption{Cross-model replication of the grounding analysis. Accuracy on the categorical cells where the DeepSeek-V4 Pro Full-regime trace cites follow-up evidence (grounded, $n$=396) versus cells where it cites only core evidence (core-only, $n$=386), for the same cells under every predictor. Participant-cluster bootstrap CIs.}
\label{tab:e19}
\end{table}

To check that the grounding results are not specific to the predictor, we re-ran the \full{} and \coreten{} regimes with \xModelName{}, an open-weight model from a different family, with the same prompts, the same two-call structure, the same per-cohort question wording and the model's default thinking mode (its API has no medium effort level); its zero-context baseline was drawn with the same model (\xNoPersonaDraws{} draws per cohort wording). Accuracy was \xFullAcc{} (95\% CI \xFullCI) with the full transcript and \xCoreAcc{} (\xCoreCI) with the core block, against \xNoPersonaAcc{} (\xNoPersonaCI) for its zero-context draws and \accMajority{} for the majority answer; the paired differences are \xFullCoreDiff{} (\xFullCoreCI, $p = \xFullCoreP$) for \full{} versus \coreten{}, \xFullNoPersonaDiff{} (\xFullNoPersonaCI, $p = \xFullNoPersonaP$) for \full{} versus its zero-context draws and \xFullMajorityDiff{} (\xFullMajorityCI, $p = \xFullMajorityP$) for \full{} versus the majority answer.
It cited follow-up evidence for \xFollowupSharePct\% of the categorical cells (GPT-5: \xGptFollowupSharePct\%), and the two predictors cite follow-ups on the same cells more often than chance ($\kappa = \xOverlapKappa$ over \xOverlapN{} cells). The grounding analysis takes the cells on which its \full{} trace cites follow-up evidence and compares, on those same cells, the \full{} prediction with its own \coreten{} prediction, which never saw the follow-ups: \full{} reaches \xPairedGroundedFull{} against \xPairedGroundedCore{} for \coreten{} (paired gain \xPairedGroundedGain, participant-cluster 95\% CI \xPairedGroundedGainCI; the follow-ups turned \xTransGroundedImproved{} wrong \coreten{} predictions right and \xTransGroundedWorsened{} right ones wrong, $p = \xTransGroundedP$, sign test), while on the cells that cite only core evidence the gain is \xPairedCoreOnlyGain{} (\xPairedCoreOnlyGainCI). Within the \full{} regime, the traces that cite follow-ups are \xGroundedDiff{} more accurate than those that cite core evidence only (\xGroundedAcc{} against \xCoreOnlyAcc; 95\% CI \xGroundedCI, $p = \xGroundedP$); on the same cells the majority answer gains \xSameMajorityDiff{} (\xSameMajorityCI) and GPT-5's \full{} predictions \xSameGptFullDiff{} (\xSameGptFullCI; Table~\ref{tab:e19}). For the second predictor, then, the follow-ups add a few points on the cells where it cites them and remove about as many on the cells where it does not, for no net gain (\full{} minus \coreten{} \xFullCoreDiff, $p = \xFullCoreP$); the within-regime gap between follow-up-citing and core-only traces is as small and unreliable as for GPT-5, and the transcript does not lift the predictor above either baseline.
Because the predictor's own evidence-location label is itself an LLM judgement, we also located each quoted span by exact string matching against the core and follow-up blocks, which involves no model: \xQuotesFoundPct\% of the \xQuotes{} quotations occur verbatim in the transcript, the two labellings agree on whether a trace cites follow-up evidence for \xLabelAgreeBinary{} of cells ($\kappa = \xLabelKappaBinary$), and under the string-matched labels the grounded-versus-core-only gap is \xGroundedDiffV{} (95\% CI \xGroundedCIV, $p = \xGroundedPV$).

\subsection{Personality recovery}
\label{app:personality}
\begin{table}[htbp]
\centering
\footnotesize
\setlength{\tabcolsep}{3pt}
\begin{tabular}{lccccc}
\toprule
Condition & $N$ & hit@2 & top-1 & off-by-1 & off-by-2 \\
\midrule
Cohort 1: Core-10 & 20 & 0.200 & 0.100 & 0.550 & 0.250 \\
Cohort 1: Full & 20 & 0.400 & 0.100 & 0.450 & 0.150 \\
Cohort 1: Summary & 20 & 0.500 & 0.400 & 0.450 & 0.050 \\
Cohort 2: Core-10 & 26 & 0.423 & 0.308 & 0.423 & 0.115 \\
Cohort 2: Full & 26 & 0.654 & 0.423 & 0.269 & 0.038 \\
Cohort 2: Static & 26 & 0.462 & 0.308 & 0.462 & 0.038 \\
Cohort 2: Summary & 26 & 0.462 & 0.346 & 0.423 & 0.077 \\
\bottomrule
\end{tabular}
\caption{MBTI recovery (auxiliary task). hit@2 and top-1: higher is better; off-by-$k$: share of participants whose closest predicted type differs on exactly $k$ axes.}
\label{tab:c1}
\end{table}

\begin{table}[htbp]
\centering
\footnotesize
\setlength{\tabcolsep}{3pt}
\begin{tabular}{lcccccc}
\toprule
Condition & $N$ & O & C & E & A & N \\
\midrule
Cohort 1: Core-10 & 20 & 0.800 & 0.550 & 0.400 & 0.600 & 0.650 \\
Cohort 1: Full & 20 & 0.850 & 0.850 & 0.300 & 0.800 & 0.300 \\
Cohort 1: Summary & 20 & 0.900 & 0.800 & 0.300 & 0.650 & 0.200 \\
Cohort 2: Core-10 & 26 & 0.731 & 0.692 & 0.654 & 0.423 & 0.538 \\
Cohort 2: Full & 26 & 0.846 & 0.577 & 0.692 & 0.423 & 0.538 \\
Cohort 2: Static & 26 & 0.808 & 0.654 & 0.654 & 0.423 & 0.615 \\
Cohort 2: Summary & 26 & 0.923 & 0.769 & 0.615 & 0.500 & 0.538 \\
\bottomrule
\end{tabular}
\caption{Big Five binarized (low/high at 21 on the 1--40 scale) match rate per dimension (auxiliary task).}
\label{tab:c2}
\end{table}

Table~\ref{tab:c1} gives the MBTI metrics per cohort and regime (hit@2, exact top-1 match, and the share of participants whose closest predicted type differs from their self-report on exactly one or two axes) and Table~\ref{tab:c2} the per-trait Big Five match rates after binarising scores at 21. Openness is the trait recovered most reliably; the weakest trait differs by cohort (extraversion and neuroticism in cohort 1, agreeableness in cohort 2).

\subsection{Instrument consistency}
\label{app:consistency}
\begin{table}[htbp]
\centering
\footnotesize
\setlength{\tabcolsep}{4pt}
\begin{tabular}{lccc}
\toprule
Scope & Within & Between & Centroid acc. \\
\midrule
Pooled & 0.606 & 0.341 & 0.991 \\
Cohort 1 & 0.636 & 0.348 & -- \\
Cohort 2 & 0.639 & 0.342 & -- \\
Cross-cohort & 0.572 & 0.339 & -- \\
\bottomrule
\end{tabular}
\caption{Semantic consistency of the per-participant generated core questions (text-embedding-3-large): mean pairwise cosine between different participants' questions for the same domain vs for different domains, and leave-one-out nearest-centroid domain accuracy (chance 0.10).}
\label{tab:e8}
\end{table}

Core questions were embedded with a sentence-embedding model (\hNQuestions{} questions from \hNParticipants{} participants; two participants whose core block could not be split into ten questions were excluded). Table~\ref{tab:e8} reports mean cosine similarity within and between domains and the leave-one-out nearest-centroid accuracy; \hParticipantsPerfect{} of \hNParticipants{} participants had all ten questions assigned to the intended domain.

\subsection{Information quantity and accuracy}
\label{app:quantity}
\begin{figure}[h]
\centering
\includegraphics[width=\linewidth]{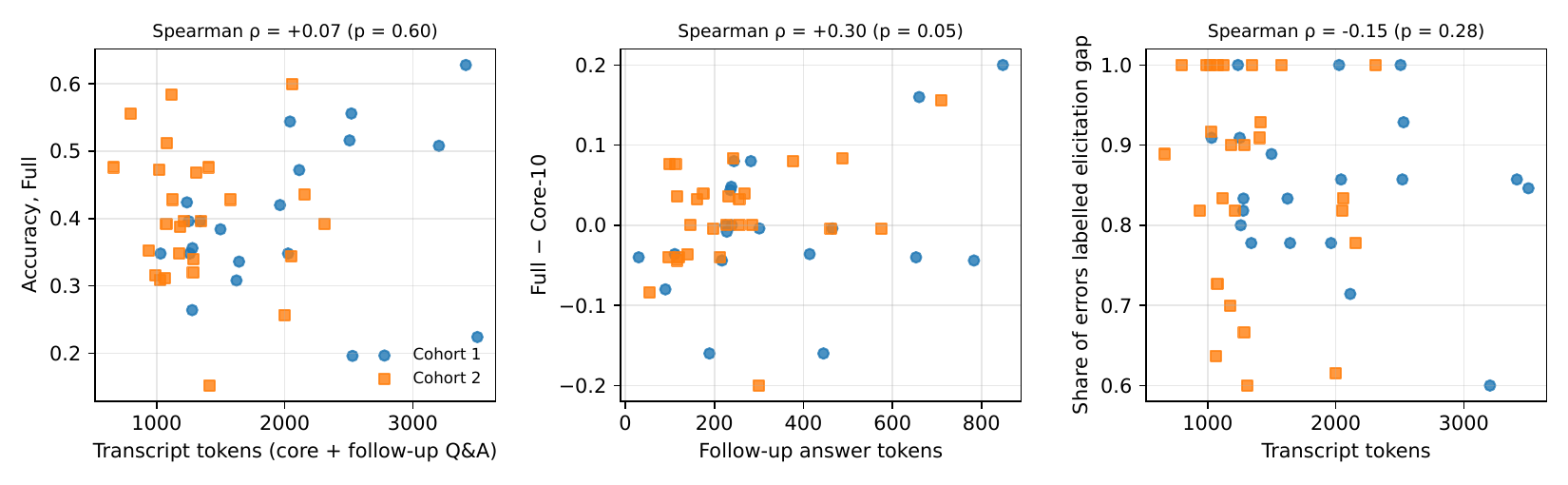}
\caption{Participant-level transcript length against \full{} accuracy, follow-up answer length against the \full{} $-$ \coreten{} lift, and transcript length against the share of a participant's errors labelled elicitation gap. Spearman $\rho$ with within-cohort permutation $p$.}
\label{fig:e14}
\end{figure}
Transcript length is unrelated to accuracy under every regime (\full: $\rho = \iTokensFullRho$, $p = \iTokensFullP$; \coreten: $\rho = \iTokensCoreRho$, $p = \iTokensCoreP$; \summ: $\rho = \iTokensSummaryRho$, $p = \iTokensSummaryP$). Follow-up answer length is weakly related to the lift of \full{} over \coreten{} (within cohort $\rho = \iFollowupTokensLiftRhoWithin$, $p = \iFollowupTokensLiftPWithin$), which is consistent with follow-ups changing some predictions without making them more accurate on average.

\subsection{Audit detail}
\label{app:audit-detail}
\begin{figure}[h]
\centering
\includegraphics[width=\linewidth]{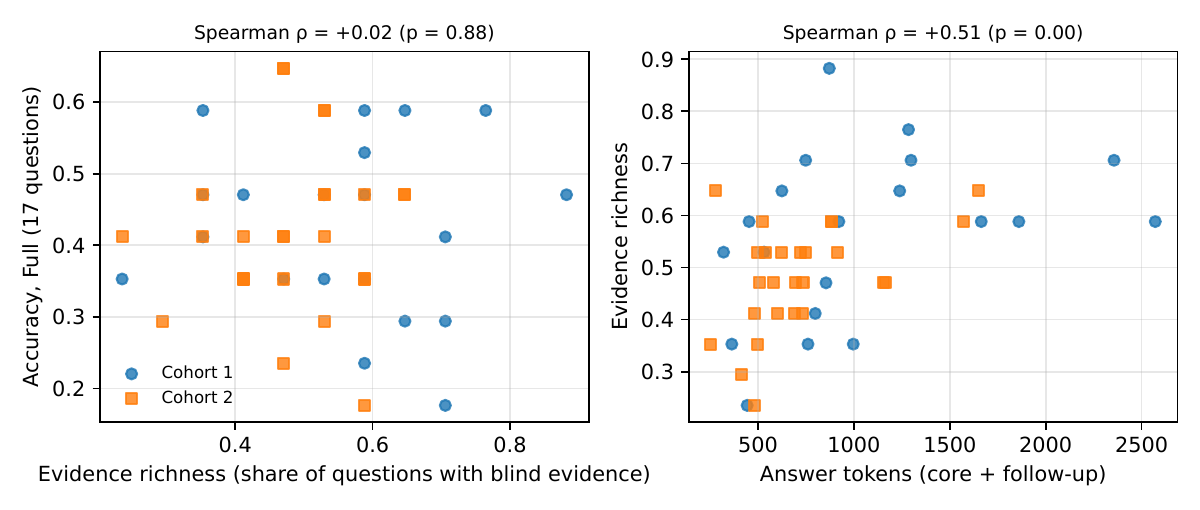}
\caption{Blind evidence audit at the participant level: evidence richness (share of the \nChoiceItems{} categorical questions with blind evidence) against \full{} accuracy on those items, and transcript length against richness.}
\label{fig:e15}
\end{figure}

Figure~\ref{fig:e15} shows the participant-level relation between evidence richness, transcript length and \full{} accuracy reported in \S\ref{sec:results-audit}.
\begin{table}[htbp]
\centering
\footnotesize
\setlength{\tabcolsep}{4pt}
\begin{tabular}{p{0.5\linewidth}r}
\toprule
Quantity & Value \\
\midrule
\textit{Blind audit (E15), 782 cells} &  \\
evidence exists: raw agreement & 0.815 \\
evidence exists: Cohen's $\kappa$ & 0.630 \\
evidence exists: Krippendorff's $\alpha$ & 0.629 \\
strength 0--3: quadratic-weighted $\kappa$ & 0.724 \\
strength 0--3: Spearman $\rho$ & 0.710 \\
favoured option agreement (both found evidence) & 0.835 ($n$=309) \\
coverage under the second annotator & 0.459 \\
precision under the second annotator & 0.343 \\
Full acc.\ with vs without evidence (second annotator) & 0.448 vs 0.397 ($p$=0.153) \\
\textit{Informed decomposition (E11), 513 traces} &  \\
four-way label: raw agreement & 0.905 \\
four-way label: Cohen's $\kappa$ & 0.579 \\
four-way label: Krippendorff's $\alpha$ & 0.578 \\
elicitation gap vs rest: $\kappa$ & 0.664 \\
evidence exists (step 1): agreement & 0.927 \\
control traces: agreement & 0.850 \\
elicitation-gap share under the second annotator & 0.905 [0.879, 0.929] \\
\bottomrule
\end{tabular}
\caption{Agreement between the first annotator (Claude Opus 5.5) and the second annotator (codex; a different model family) over the same packets and protocols, with the headline quantities recomputed under the second annotator.}
\label{tab:agreement_codex}
\end{table}

Table~\ref{tab:agreement_codex} reports agreement between the two annotators and the headline quantities recomputed under the second one. Disagreements on the informed labels concentrate in the three small classes: \kCodexBShareOfARetPct\% of the first annotator's retrieval labels, \kCodexBShareOfCIntPct\% of its integration labels and \kCodexBShareOfAReasPct\% of its reasoning labels become elicitation gap under the second annotator, which is why its elicitation-gap share is higher. Under the second annotator, \full{} accuracy with versus without blind evidence is \kCodexBlindFullDiff{} ($p = \kCodexBlindFullP$). \begin{table}[htbp]
\centering
\footnotesize
\setlength{\tabcolsep}{3pt}
\begin{tabular}{lccccc}
\toprule
Group & $n$ & B elicit. & A retr. & C integr. & A reas. \\
\midrule
\multicolumn{6}{l}{\textit{By cohort}} \\
cohort 1 & 196 & 0.85 & 0.03 & 0.04 & 0.08 \\
cohort 2 & 257 & 0.85 & 0.04 & 0.05 & 0.06 \\
\multicolumn{6}{l}{\textit{By the predictor's evidence-location label}} \\
both blocks & 109 & 0.81 & 0.06 & 0.05 & 0.09 \\
core block & 267 & 0.86 & 0.04 & 0.05 & 0.05 \\
follow-up block & 58 & 0.83 & 0.02 & 0.03 & 0.12 \\
no evidence cited & 19 & 1.00 & 0.00 & 0.00 & 0.00 \\
\multicolumn{6}{l}{\textit{By the predictor's reasoning-type label}} \\
constraint & 27 & 0.81 & 0.04 & 0.04 & 0.11 \\
generic & 9 & 1.00 & 0.00 & 0.00 & 0.00 \\
narrative & 21 & 0.76 & 0.00 & 0.14 & 0.10 \\
value & 396 & 0.85 & 0.04 & 0.04 & 0.07 \\
\bottomrule
\end{tabular}
\caption{Informed error labels (shares within group) by cohort, by the predictor's own evidence-location label, and by its reasoning-type label.}
\label{tab:e11_breakdown}
\end{table}

Table~\ref{tab:e11_breakdown} breaks the informed labels down by cohort and by the predictor's own evidence-location and reasoning-type labels; traces for which the predictor cited no evidence are all elicitation gaps, and reasoning failures are most common among follow-up-grounded traces. The predictor's own evidence-location and reasoning-type labels were checked by humans on cohort 1: \hvAnnotators{} annotators verified \hvTraces{} traces (\hvOverlap{} by all three; inter-rater agreement \hvIRAPct\%) and agreed with the predictor's labels in \hvAgreePct\% of cases.
Blind evidence strength was 1 (an indirect disposition) in \fStrengthOneN{} cells, 2 (a clearly applicable value or habit) in \fStrengthTwoN{} and 3 (an explicit statement about the kind of situation) in \fStrengthThreeN{}; within the \full{} regime, accuracy by strength is \fFullAccStrengthZero{} (none), \fFullAccStrengthOne{} (indirect), \fFullAccStrengthTwo{} (moderate) and \fFullAccStrengthThree{} (explicit). A reader who answers every cell from the transcript alone, with cells lacking evidence scored as misses, reaches \fReaderAcc{} on the categorical questions. One participant's first blind pass was discarded because the annotator had inadvertently seen the participant's answers; it was redone from scratch in a fresh context. The informed auditor's confidence was 2 or lower for \eConfLowPct\% of wrong traces, and \eFlagInconsistent{} traces were flagged as cases in which the participant's answer contradicted their own stated values.

\subsection{Sensitivity to the three rewritten transcripts}
\label{app:sensitivity}
\begin{table*}[tbp]
\centering
\scriptsize
\setlength{\tabcolsep}{3pt}
\begin{tabular}{p{4.3cm}cc}
\toprule
Quantity & All participants & Without c1-18, c2-11, c2-20 \\
\midrule
Accuracy: Cohort 1, No-persona & 0.427 [0.390, 0.463] & 0.425 [0.386, 0.462] \\
Accuracy: Cohort 1, Core-10 & 0.401 [0.359, 0.444] & 0.408 [0.367, 0.450] \\
Accuracy: Cohort 1, Full & 0.399 [0.351, 0.450] & 0.408 [0.362, 0.459] \\
Accuracy: Cohort 1, Summary & 0.393 [0.350, 0.433] & 0.398 [0.356, 0.439] \\
Accuracy: Pooled, Core-10 & 0.395 [0.365, 0.429] & 0.402 [0.367, 0.432] \\
Accuracy: Pooled, Full & 0.400 [0.373, 0.431] & 0.406 [0.376, 0.437] \\
Accuracy: Pooled, Summary & 0.388 [0.357, 0.421] & 0.386 [0.351, 0.420] \\
Accuracy: Pooled, No-persona & 0.392 [0.356, 0.425] & 0.389 [0.352, 0.423] \\
Cohort 1: grounded vs core-only acc. & 0.455 (134) vs 0.393 (201) & 0.462 (130) vs 0.404 (188) \\
Cohort 1: transitions improved / worsened & 15 / 6 & 15 / 6 \\
Pooled: grounded vs core-only acc. & 0.445 (301) vs 0.397 (443) & 0.454 (282) vs 0.397 (416) \\
Pooled: transitions improved / worsened & 32 / 21 & 30 / 20 \\
Contrast 5: grounded $-$ core-only [95\% CI] & +0.048 [-0.033, +0.131] & +0.057 [-0.027, +0.145] \\
Contrast 5: permutation $p$ & 0.250 & 0.151 \\
Contrast 3: Full $-$ Core-10 (pooled) [95\% CI] & +0.005 [-0.017, +0.027] & +0.004 [-0.019, +0.028] \\
Contrast 3: permutation $p$ & 0.665 & 0.716 \\
\bottomrule
\end{tabular}
\caption{Sensitivity to the three participants whose core answers are not in their own words (c1-18 dictated answers that an LLM condensed; c2-11's core answers are an LLM restatement in the second person; c2-20 submitted LLM-written summaries): main accuracies (participant-bootstrap 95\% CIs) and the evidence-use contrasts with and without them.}
\label{tab:c118}
\end{table*}

Table~\ref{tab:c118} recomputes the main quantities without the three participants whose core answers were rewritten by an LLM before reaching us (c1-18, c2-11, c2-20); no conclusion changes.

\subsection{Default biases among reasoning failures}
\label{app:biases}
The \eAReasN{} reasoning failures identified by the informed audit show three recurring defaults. \emph{Normative rule bias}: in a legal scenario with insufficient evidence (Q21) the predictor consistently chose strict procedural adherence, releasing the suspect and continuing the investigation, where participants chose more cautious or situational responses; in Q25 it prioritised informational completeness and procedural fairness where participants chose by practical trade-off. \emph{Prosocial harmony bias}: asked how one responds to a friend who repeatedly cancels (Q2), the predictor chose communication and repair where participants chose to disengage or set a boundary; in Q22 it chose inclusive activities that accommodate everyone. \emph{Over-responsibility bias}: in Q24 the predictor chose ``two-win'' solutions that honour every obligation where participants chose a single pragmatic option under real constraints. These defaults match documented tendencies of aligned models toward norm-consistent, cooperative and responsibility-sensitive answers \citep{ouyang2022training,sharma2024sycophancy,scherrer2023evaluating,cheung2025amplified,kim2026relational,dash2025motivated}.

\FloatBarrier
\section{Related work: design comparison}
\label{app:related-table}
\begin{table*}[t]
\centering\footnotesize\setlength{\tabcolsep}{3pt}
\begin{tabular}{p{2.9cm}p{3.3cm}p{1.0cm}p{3.0cm}p{2.4cm}p{1.8cm}}
\toprule
Study & Elicitation & $N$ & Individual-level target & Adaptive vs fixed control & Audit of evidence \\
\midrule
\citet{park2024generative} & two-hour AI-led interview & 1,052 & General Social Survey items, experiments & none (deletion and summary ablations) & none \\
\citet{deng2026aimoderated} & AI-moderated interview vs static questionnaire & 139 / 154 & marketing ratings and rankings & between-subjects; static arm without follow-ups & none \\
\citet{li2025hugagent} & LLM-led interview & 54 & within-domain belief updates & none & none (single-letter outputs) \\
\citet{wang2026interview} & interviews turned into agents & -- & product-concept ratings & none & none \\
\citet{peng2025funhouse}; \citet{toubia2025twin} & survey-panel personas & 2,000+ & outcomes of 19 studies & -- & none \\
\citet{ahn2026itemmean}; \citet{kinzinger2026synthetic}; \citet{ye2026beyond}; \citet{pourtaheri2026behavioral} & persona-representation ablations on panel or history data & -- & survey items, retest outcomes & -- & none \\
\citet{chopra2026qualitative}; \citet{austin2026experimental} & AI interview vs fixed questions & -- & data quality (coverage, words), not prediction & between-subjects & none \\
\citet{li2023gate} & LLM-led preference elicitation vs prompts & -- & preferences incl.\ moral vignettes & time-matched arms & none \\
\citet{lee2025solar} & years of public comments & -- & individual moral judgements & -- & none \\
This paper & one-hour LLM-led interview & \nParticipants{} & \nItems{} value-laden dilemma decisions; MBTI, Big Five & within-subject: fixed vs adaptive follow-ups, same core questions & informed decomposition and blind evidence audit; two annotator families \\
\bottomrule
\end{tabular}
\caption{Design comparison with the closest studies of interview- or persona-based simulation of individuals (identifiers verified against the publishers' or arXiv records; details in \S\ref{sec:related}). ``--'': not applicable or not reported in the source we consulted.}
\label{tab:related}
\end{table*}
Table~\ref{tab:related} places this study among the works discussed in \S\ref{sec:related} along the four dimensions that matter for our question: how the persona was elicited, whether the target is an individual-level decision, whether adaptive questioning was compared with a fixed alternative on the same people, and whether anyone examined what the elicited material contained.

\FloatBarrier
\section{Participants and data release}
\label{app:demographics}
\begin{table}[htbp]
\centering
\footnotesize
\setlength{\tabcolsep}{3pt}
\begin{tabular}{llr}
\toprule
 & Level & $N$ \\
\midrule
\textit{Age band} &  &  \\
 & under 20 & 5 \\
 & 20-30 & 36 \\
 & 30-40 & 5 \\
\textit{Gender} &  &  \\
 & women & 28 \\
 & men & 18 \\
\textit{Residence} &  &  \\
 & North America & 24 \\
 & China & 22 \\
\bottomrule
\end{tabular}
\caption{Participant demographics, both cohorts combined, from the recruitment records (aggregate counts only; the records cannot be linked to individual transcripts and are not part of the release).}
\label{tab:demographics}
\end{table}

Table~\ref{tab:demographics} gives the aggregate counts from the recruitment records (recruitment in \S\ref{sec:participants}). The study was fully anonymous, so these counts cannot be linked to individual transcripts and are not released with them.

\paragraph{Release.}
\label{sec:data}
We release, for participants who consented: de-identified interview transcripts (core questions and answers, adaptive follow-ups, fixed follow-ups, summaries), the \nItems{} questions, each participant's answers and personality self-reports, every prediction with its trace under every regime, the zero-context draws, the informed-audit labels with quoted spans, the blind-audit labels, both audit protocols, and the analysis code that regenerates every table and number in this paper from the released files. During review the released files accompany the submission as anonymised supplementary material; the public repository is linked on acceptance.

\FloatBarrier
\section{Vignette: a model with long-term memory of one participant}
\label{app:vignette}
One author (participant c1-01 in cohort 1) asked a general-purpose assistant that had accumulated months of persistent memory about them to predict their \nItems{} answers. Its accuracy was \vChatgptAcc, against \vFullAcc{} for the \full{} regime, \vCoreAcc{} for \coreten, \vNoPersonaAcc{} for the zero-context predictor and \vMajorityAcc{} for the majority answer on the same participant (the \full{} regime ranks them \vRank{} of \vRankOf{} cohort-1 participants, so they are the best-predicted participant in that cohort). The memory-based predictions agreed with the \full{} predictions on \vSameAsFull{} of \nItems{} questions. The assistant's rationales show that it had encountered the study materials, so its memory is not independent of this project; the vignette is $n = 1$ and illustrative only. It does suggest that long exposure to a person's conversation does not by itself supply the situated decision policies the dilemmas require.

\end{document}